\documentclass{article}

\usepackage{arxiv}

\usepackage[utf8]{inputenc} % allow utf-8 input
\usepackage[T1]{fontenc}    % use 8-bit T1 fonts
\usepackage{hyperref}       % hyperlinks
\usepackage{url}            % simple URL typesetting
\usepackage{booktabs}       % professional-quality tables
\usepackage{amsfonts}       % blackboard math symbols
\usepackage{nicefrac}       % compact symbols for 1/2, etc.
\usepackage{microtype}      % microtypography
\usepackage{graphicx}
\usepackage{natbib}
\usepackage{doi}

\usepackage[ruled,vlined]{algorithm2e}
\usepackage{siunitx}  % Aligning numbers by decimal points in table columns
\usepackage{etoolbox}
\usepackage{fancyvrb}
\usepackage{amsmath}
\usepackage{float}

\usepackage{todonotes}
\usepackage{comment}

\newcommand{\rnc}{\texttt{RNC}}
\newcommand{\drnc}[1]{\texttt{DRNC(#1)}}
\newcommand{\ripper}{\texttt{Ripper}}
\newcommand{\CART}{\texttt{CART}}

\graphicspath{{figures/}}

\title{An Empirical Investigation into Deep and Shallow Rule Learning}

\author{ \href{https://orcid.org/0000-0003-3183-2953}{\includegraphics[scale=0.06]{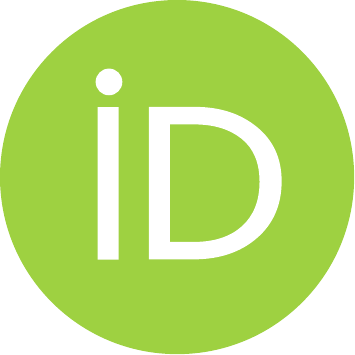}\hspace{1mm}Florian Beck} \\
	Application-oriented Knowledge Processing (FAW) \\
	Department of Computer Science \\
	Johannes Kepler University Linz, Austria \\
	\texttt{fbeck@faw.jku.at} \\
	\And
	\href{https://orcid.org/0000-0002-1207-0159}{\includegraphics[scale=0.06]{orcid.pdf}\hspace{1mm}Johannes F\"urnkranz} \\
	Application-oriented Knowledge Processing (FAW) \\
	Department of Computer Science \\
	Johannes Kepler University Linz, Austria \\
	\texttt{juffi@faw.jku.at} \\
}

\date{}

\renewcommand{\headeright}{}
\renewcommand{\undertitle}{}
\renewcommand{\shorttitle}{Deep and Shallow Rule Learning}

\hypersetup{
pdftitle={An Empirical Investigation into Deep and Shallow Rule Learning},
pdfsubject={cs.LG},
pdfauthor={Florian Beck, Johannes F\"urnkranz},
pdfkeywords={Inductive rule learning, Deep learning, Learning in logic, Mini-batch learning, Stochastic optimization},
}

\begin{document}
\maketitle

\begin{abstract}
Inductive rule learning is arguably among the most traditional paradigms in machine learning. Although we have seen considerable progress over the years in learning rule-based theories, all state-of-the-art learners still learn descriptions that directly relate the input features to the target concept. In the simplest case, concept learning, this is a disjunctive normal form (DNF) description of the positive class. While it is clear that this is sufficient from a logical point of view because every logical expression can be reduced to an equivalent DNF expression, it could nevertheless be the case that more structured representations, which form deep theories by forming intermediate concepts, could be easier to learn, in very much the same way as deep neural networks are able to outperform shallow networks, even though the latter are also universal function approximators. In this paper, we empirically compare deep and shallow rule learning with a uniform general algorithm, which relies on greedy mini-batch based optimization.
Our experiments on both artificial and real-world benchmark data indicate that deep rule networks outperform shallow networks.
\end{abstract}

% keywords can be removed

\section{Introduction}

Dating back to the AQ algorithm \citep{AQ}, inductive rule learning is one of the most traditional fields in inductive rule learning. However, when reflecting upon its long history \citep{jf:Book-Nada}, it can be argued that 
while modern methods are somewhat more scalable than traditional rule learning algorithms \citep[see, e.g.,][]{RuleSets-Bayesian,InterpretableDecisionSets}, no major break-through has been made. In fact, the \ripper rule learning algorithm \citep{Ripper} is still very hard to beat in terms of both accuracy and simplicity of the learned rule sets. 
All these algorithms, traditional or modern, typically provide flat lists or sets of rules, which directly relate the input variables
to the desired output. In the simplest setting, concept learning, where the goal is to learn a set of rules that collectively describe the target concept, the learned set of rules can be considered as a logical expression in disjunctive normal form (DNF), in which each conjunction forms a rule that predicts the positive class.

In this paper, we argue that one of the key factors for the strength of 
deep learning algorithms is that latent variables are formed during the learning
process. However, while neural networks excel in implementing this ability in their hidden layers, which can be effectively trained via backpropagation, there is essentially no counter-part to this ability in inductive rule learning.
We therefore set out to verify the hypothesis that deep rule structures might be easier to learn than flat rule sets, in very much the same way as deep neural networks have a better performance than single-layer networks. Note that this is not obvious, because, in principle, every logical formula can be represented with a DNF expression, which corresponds to a flat rule set. Our tool of investigation is a simple stochastic optimization algorithm to optimize a rule network of a given size. While this does not quite reach state-of-the-art performance (in either setting, shallow or deep), it nevertheless allows us to gain some insights into these settings. We also test on both, real-world UCI benchmark datasets, as well as artificial datasets for which we know the underlying target concept representations.

The remainder of the paper is organized as follows: Sect.~\ref{sec:deep-rule-learning} elaborates why deep rule learning is of particular interest and refers to related work. We propose a new network approach in Sect.~\ref{sec:deep-rule-networks} and test it in Sect.~\ref{sec:experiments}. The results are concluded in Sect.~\ref{sec:conclusion}, followed by possible future extensions and improvements in Sect.~\ref{sec:future-work}.

\section{Deep Rule Learning}
\label{sec:deep-rule-learning}

In this section, we will briefly discuss the state-of-the-art in learning deep, structured rule bases. We start with a brief motivation, and continue to review related work in several relevant areas, including constructive induction, multi-label rule learning, or binary and ternary networks.

\subsection{Motivation}
\label{sec:motivation}

Rule learning algorithms typically provide flat lists that directly relate the input
to the output. Consider, e.g., the following example:
%, taken from \citet{jf:MLJ-CognitiveBias}:  
the parity concept, which is known to be hard to learn for heuristic, greedy learning algorithms, checks whether an odd or an even number of $R$ relevant attributes (out of a possibly higher total number of attributes) are set to \texttt{true}.
Figure~\ref{fig:parity}a shows a flat rule-based representation\footnote{We use a Prolog-like notation for rules, where the consequent (the head of the rule) is written on the left and the antecedent (the body) is written on the right. For example, the first rule 
%in Figure~\ref{fig:flat-parity} 
reads as: If x1, x2, x3 and x4 are all true and x5 is false then parity holds.} of the target concept for $R = 5$, which requires $2^{R-1} = 16$ rules. On the other hand, a structured representation, which introduces three auxiliary predicates (\texttt{parity2345}, \texttt{parity345} and \texttt{parity45} as shown in Figure~\ref{fig:parity}b), is much more concise using only $2\cdot(R-1) = 8$ rules. We argue that the parsimonious structure of the latter could be easier to learn because it uses only a linear number of rules, and slowly builds up the complex target concept \texttt{parity} from the smaller subconcepts \texttt{parity2345}, \texttt{parity345} and \texttt{parity45}. 
%This is in line with the criticism of \citet{FuzzyRules-DataDriven} who argued that the flat structure of fuzzy rules is one of the main limitations of current fuzzy rule learning systems.

\begin{figure*}[h]
	
\begin{minipage}{.6\textwidth}
	\centering
	\begin{footnotesize}
		\begin{Verbatim}
parity :-     x1,     x2,     x3,     x4, not x5.
parity :-     x1,     x2, not x3, not x4, not x5.
parity :-     x1, not x2,     x3, not x4, not x5.
parity :-     x1, not x2, not x3,     x4, not x5.
parity :- not x1,     x2, not x3,     x4, not x5.
parity :- not x1,     x2,     x3, not x4, not x5.
parity :- not x1, not x2,     x3,     x4, not x5.
parity :- not x1, not x2, not x2, not x4, not x5.
parity :-     x1,     x2,     x3, not x4,     x5.
parity :-     x1,     x2, not x3,     x4,     x5.
parity :-     x1, not x2,     x3,     x4,     x5.
parity :- not x1,     x2,     x3,     x4,     x5.
parity :- not x1, not x2, not x3,     x4,     x5.
parity :- not x1, not x2,     x3, not x4,     x5.
parity :- not x1,     x2, not x3, not x4,     x5.
parity :-     x1, not x2, not x2, not x4,     x5.
		\end{Verbatim}
		\end{footnotesize}
		
%		\subcaption{A flat unstructured rule set for the parity concept}
\medskip
(a) A flat unstructured rule set for the parity concept
%		\label{fig:flat-parity}
\end{minipage}%		
\begin{minipage}{.39\textwidth}
	\centering		
	\begin{footnotesize}
\begin{Verbatim}


parity45   :-     x4,     x5.
parity45   :- not x4, not x5.
	
parity345  :-     x3, not parity45.
parity345  :- not x3,     parity45.  
	
parity2345 :-     x2, not parity345.
parity2345 :- not x2,     parity345.   

parity     :-     x1, not parity2345.
parity     :- not x1,     parity2345. 



\end{Verbatim}
		
	\end{footnotesize}
	
	% \subcaption{A deep structured rule base for parity using three auxiliary predicates}
	\medskip 
(b) A deep structured rule base for parity using three auxiliary predicates
%	\label{fig:deep-parity}
\end{minipage}	
	
\caption{Unstructured and structured rule sets for the parity concept.}
\label{fig:parity}
\end{figure*}

%However, while the deep rule set is obviously more compact, it is not clear how such deep structures can be learned, or whether they offer any advantages in predictive performance. We are convinced that structured rule sets, once we are able to learn them, offer advantages in both, learnability and predictive accuracy, but this needs to be thoroughly investigated and assessed in experimental evaluation.

\begin{figure}[t]
\begin{minipage}[b]{0.39\textwidth}
	\centering
\includegraphics[angle=270,width=38mm]{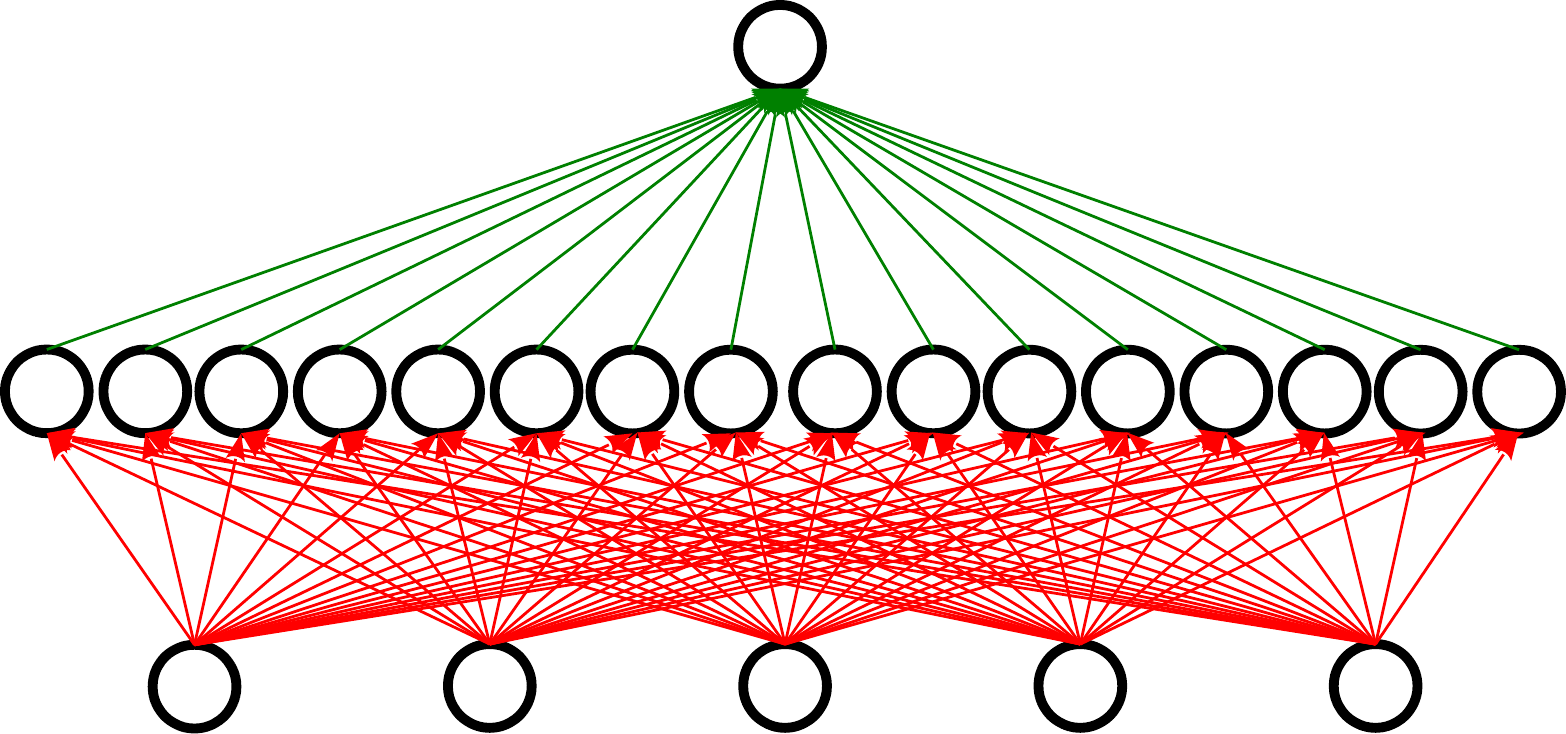}
%\caption{Flat representation (cf.\ Figure~\ref{fig:flat-parity})}
%\label{fig:flat-parity-network}

\medskip
(a) shallow representation
\end{minipage}%
\begin{minipage}[b]{0.6\textwidth}
	\centering
\includegraphics[angle=270,width=90mm]{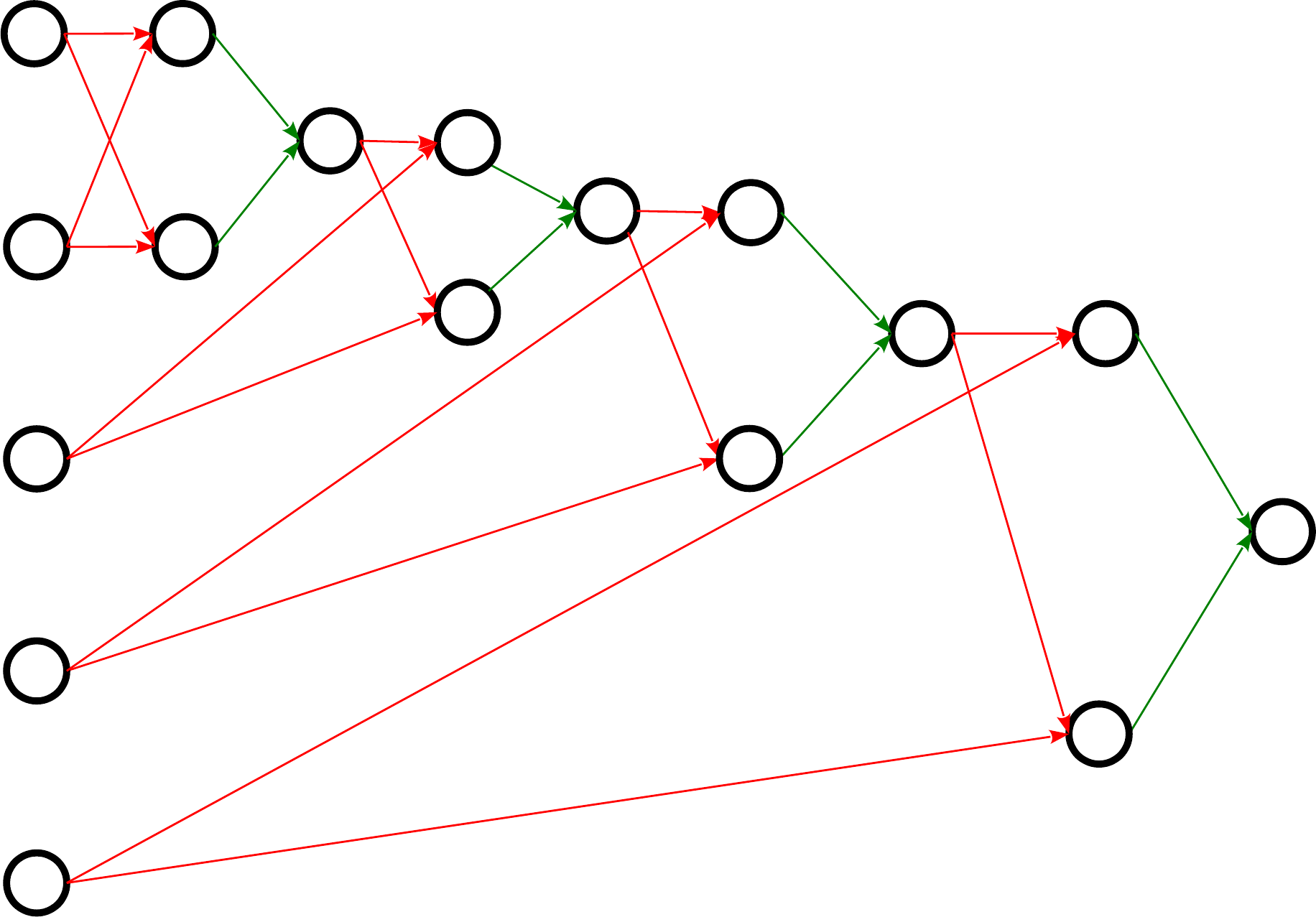}

\vspace*{10mm}
\medskip
(b) deep representation
%\caption{Deep representation (cf.\ Figure~\ref{fig:deep-parity})}
%\label{fig:deep-parity-network}
\end{minipage}
\caption{Network representations of the parity rule sets of Figure~\ref{fig:parity}. Red connections are logical ANDs, green edges correspond to logical ORs.}
\label{fig:parity-network}
\end{figure}

To motivate this,
% For the moment, 
we draw an analogy to neural network learning, and view 
rule sets as networks. Conventional rule learning algorithms learn a flat rule set of the type shown in Figure~\ref{fig:parity}a, which may be viewed as a concept description in disjunctive normal form (DNF): Each rule body corresponds to a single conjunct, and these conjuncts are connected via a disjunction (each positive example must be covered by one or more of these rule bodies). This situation is illustrated in Figure~\ref{fig:parity-network}a, where the 5 input nodes are connected to 16 hidden nodes - one for each of the 16 rules that define the concept - and these are then connected to a single output node. Analogously, the deep parity rule set of Figure~\ref{fig:parity}b may be encoded into a deeper network structure as shown in Figure~\ref{fig:parity-network}b. 
Clearly, the deep network is more compact and considerably sparser in the number of edges. Of course, we need to take into consideration that the optimal structure is not known beforehand and presumably needs to emerge from a fixed network structure that offers the possibility for some redundancy, but nevertheless we expect that such structured representations offer similar advantages as deep neural networks offer over single-layer networks.

It is important to note that deep structures do not increase the expressiveness of the learned concepts. Any formula in propositional logic (and we limit ourselves to propositional logic in this project) can be converted to a DNF formula. In the worst case (a so-called \emph{full DNF}), each of the input variables appears exactly once in all of the inputs, which essentially corresponds to enumerating all the positive examples. Thus, the size of the number of conjuncts in a DNF encoding of the inputs may grow exponentially with the number of input features. This is in many ways analogous to the universal approximation theorem \citep{NN-UniversalApproximation}, which essentially states that any continuous function can be approximated arbitrarily closely with a shallow neural network with a single hidden layer, provided that the size of this layer is not bounded.  So, in principle, deep neural networks are not necessary, and indeed, much of the neural network research in the 90s has concentrated on learning such two-layer networks. Nevertheless, we have now seen that deep neural networks are easier to train and often yield better performance, presumably because they require exponentially less parameters than shallow networks \citep{NN-WhyDeep}. In the same way, we expect that deep logical structures will yield more efficient representations of the captured knowledge and might be easier to learn than flat DNF rule sets.

\subsection{State-of-the-Art in Deep Rule Learning}

As mentioned above, the problem of deep rule learning has only rarely been explicitly addressed in the literature. 
Modern rule learning algorithms rely on ensemble-based sequential loss minimization.
 \citet{RuleFit}, e.g., learn a sparse linear model from features that have been obtained from the rules corresponding to the leaves of a decision tree ensemble 
such as a
random forest \citep{RandomForests}. 
%This essentially corresponds to a incremental layerwise forward induction of a shallow network of the type shown in Figure~\ref{fig:flat-parity-network}. 
Algorithms like ENDER \citep{ENDER} or BOOMER \citep{mr:ECML-PKDD-20} 
integrate the rule induction into the boosting procedure by aiming at the minimization of an overall regularized loss function during the learning of individual rules.
The learning algorithm for finding interpretable decision sets \citep{InterpretableDecisionSets} explicitly includes several biases for interpretability into the objective function and proposes smooth stochastic search, a method for efficiently finding an approximative solution.
\citet{RuleLists-Optimal} demonstrate an algorithm that is able to find exact loss minimizing rules.

All these algorithms are single-concept learners, i.e., they learn rules for a single target concept. However, as has been argued by \citet{jf:DeCoDeML-20-Challenge}, works in several related areas are quite relevant to the problem. In the following, we briefly review approaches that are able to autonomously discover hidden, auxiliary concepts (Section~\ref{sec:constructive-induction}), or to learn multiple dependent target concepts (Section~\ref{sec:multi-target}), and even review a few algorithms that learn logical networks (Section~\ref{sec:logic-networks}).

\subsubsection{Learning Intermediate Concepts}
\label{sec:constructive-induction}

For the general case of learning structured declarative rule sets, a key challenge is how to define and train the intermediate, hidden concepts $h_i$ which may be used for improving the final prediction. Note that in a conventional, flat structure as in Figure~\ref{fig:parity-network}a, all $h_i$ always had a fairly clear semantic in that they capture some aspect of the target variable $y$. The rule that predicts $h_i$ essentially defines a local pattern for $y$ \citep{jf:Dagstuhl-04}. 

However, when learning deeper structures, other hidden concepts need to be defined which do not directly correspond to the target variable, as can be seen from the structured parity concept of Figure~\ref{fig:parity}b. 
%Although there are machine learning systems that can tackle simple problems, 
% like the family domain, 
%there is no
%In their experiments, \citet{PI-Comprehensibility} used manually constructed logic programs for the toy domain of family relations, which is frequently used as introductory examples in logic programming.  algorithms can construct
%rules in this domain, 
%In fact, research in machine learning has not yet produced a 
%system that is powerful enough to learn deeply structured logic theories for realistic problems.
%, on which we could rely for experimentally testing this hypothesis. In machine learning, this 
This line of work has been known as
\emph{constructive induction} \citep{CI-Framework} or \emph{predicate invention}
\citep{ILP-PI},
but surprisingly, it has not received much attention since the classical works in inductive logic programming in the 1980s and 1990s.
One approach is to use a wrapper to scan for regularly co-occurring patterns in rules, and use them to define new intermediate concepts which allow to compress the original theory \citep{AQ17-HCI,CiPF}. %,New-MDL}. 
Alternatively, one can directly
invoke so-called predicate invention operators during the learning process, as, e.g., in Duce \citep{Duce}, which operates in propositional logic, and its successor systems in first-order logic \citep{CIGOL,CHAMP,PI-Statistical}. Similar to Duce, systems like MOBAL \citep{Mobal} not only try to learn theories from data, but also provide functionalities for reformulating and restructuring the knowledge base \citep{Sommer-Diss}.
More recently, \citet{ILP-PI-MetaInterpretative} introduce a technique that employs user-provided meta rules for proposing new predicates, which allow it to invent useful predicates from only very training examples. \citet{HigherLevelRepresentations} provides an excellent recent summary of work in this area.

\subsubsection{Learning Multiple Dependent Concepts}
\label{sec:multi-target}

Much of the work in machine learning is devoted to single prediction tasks, i.e., to tasks where an input vector is mapped to a single output value. When aiming to learn a deep rule base, however, one has to tackle the problem of learning a network of multiple, possibly mutually dependent concepts. 
%Pioneering works in this area can again be found in the areas of relational learning and inductive logic programming. For example, \citet{MPL} discussed the problem of multiple predicate learning in general, while also considering recursive and mutually recursive predicates which can not be organized in a DAG.
A pioneering work in this area is \citet{LabelDepRuleLearning}, which gives a broad discussion of the problem of learning multiple dependent concepts in the form of a dependency graph. Back then, the problem has primarily been studied in inductive logic programming and relational learning \citep[see, e.g.,][]{MPL}, but it has recently reappeared in multilabel classification \citep{Multilabel-Overview,tsoumakas10MLoverview,zhang2014review} and, more generally, in multi-target prediction \citep{MultiTargetPrediction}.

In fact, most of the research in multi-label classification aims for the development of methods that are capable of modeling label dependencies \citep{MultiLabel-Dependence}. One of the best-known approaches are so-called classifier chains (CC) \citep{ClassifierChains,ClassifierChains-Perspective}, which learn the labels in some (arbitrary) order where the predictions for previous labels are included as features for  subsequent models. Several extensions of this framework have been studied, such as \citet{Burkhardt2015}, who 
propose to cluster labels into  sequential blocks of sets of labels.
A general framework proposed by  \citet{Read-Hollmen-IDA-14,Read-Hollmen-arxiv-15} 
formulates multi-label classification problems as deep networks where label nodes are a special type of hidden nodes which can appear in multiple layers of the networks. 

However, while these algorithms all aim at learning multiple interconnected models, they are not capable of explicitly defining intermediate, auxiliary concepts. Some works that aim at finding so-called label embeddings \citep[e.g.,][]{jn:AAAI-16} may be viewed in this context, but they do not learn rule-based descriptions.
Rules are particularly interesting for solving this kind of problems because they allow to explicitly formalize and model dependencies between labels and between data and labels in an explicit and seamless way \citep{eh:RuleML-20}. \citet{mr:ECML-PKDD-20} propose an efficient boosting-based rule learner for multi-label classification.

\subsubsection{Discrete Deep Networks}
\label{sec:logic-networks}

Finally, in the wake of the success of deep neural networks, a few approaches have been developed that explicitly aim at learning networks with a logical structure. Sum-product networks \citep[SPNs;][]{SPNs} have an analogous structure to our AND/OR networks, but aim at modeling probability distributions instead of logical expressions.
Of particular interest to our study is the work of
\citet{SPN-DeepVsShallow}, who compare deep and shallow SPNs, and find that deep structures can result in more compact representations, which is in line with the motivation of our work. 

Somewhat closer to logic are frameworks such as TensorLog \citep{TensorLog} that aim at making probabilistic logical reasoning differentiable and therefore amenable to implementation and optimization in a deep learning environment. For example, the approach of  \citet{ILP-DeepNoisy} is able to learn logical theories from data in a matter that is considerably more robust than traditional techniques from inductive logic programming. However, it only learns to weight rules that can be generated from a set of predefined templates. In particular, no auxiliary, hidden predicates can emerge from the learner. Fuzzy pattern trees \citep{FuzzyPatternTrees} may be viewed in this way in that they build up a hierarchical structure of generalized logical functions, so that their internal nodes may be viewed as intermediate fuzzy logical concepts.

Most relevant to our work are binary networks \citep{BinaryConnect,BinaryNN-Survey}, which restrict the weights to values $\{-1,1\}$. However, their semantics does typically  not correspond to conventional logic rules, in that they enforce every feature to contribute to the function to be learned, either in its positive or negated form. Ternary networks \citep{TernaryNetworks,TernaryNetworks-Quantization}, with weights $\{-1,0,1\}$, where $0$ corresponds to ignoring the corresponding feature in the rule, could provide a solution to this, and are, indeed, quite similar in spirit to the networks we train in the remainder of this paper. Typically, they train a full deep neural network, and subsequently quantize the resulting weights to the desired two or three values, in order to allow a more compact representation and faster inference. Nevertheless, we have not made use of them in our work, because we wanted to focus on a simple optimization algorithm that is invariant for deep and shallow structures. For essentially the same reason, we have also not used state-of-the-art flat rule learning algorithms, so that observed differences in performance can be clearly attributed to differences in the network structure, and not in the optimization algorithms.

\begin{comment}

7. Yang, F., Yang, Z., Cohen, W.W.: Differentiable learning of logical
rules for knowledge base completion. CoRR abs/1702.08367 (2017),
http://arxiv.org/abs/1702.08367

\citep{TensorLog} is a differentiable deductive database focus on inference in first-order logic Program + DB + Query define a proof graph, where nodes are conjunctions of goals and edges are labeled with sets of features
uses mini-batches, which give a good speedup on small databases, but less so on larger ones.
I guess it is similar to Whirl, in that it computes similarity-based conjunction?
Task 5.4. seems to indicate this.
It is similar to Problog and to Markov logic networks, but integrates it into a deep learning infrastructure. makes it more efficient.

logic as a network architecture
Towell 

stochastic logic programs (Cussens 2001) are a predecessor of TensorLog

or to develop hybrid models that combine the interpretability of logic with the predictive strength of statistical and probabilistic models
%It has also been argued that probabilistic reasoning may not always be the best way for dealing with uncertainty and that logic reasoning may sometimes be preferable 
\citep{LogicUncertainty,DeepLogicNetworks,HarnessingDNNs}.
%\todo{There are also framework for the generic ML method like Explain (Robnik-\v{S}ikonja and Kononenko, 2008)and IME (\v{S}trumbelj and Kononenko, 2010), these could also be mentioned. }

\end{comment}

\section{Deep Rule Networks}
\label{sec:deep-rule-networks}
For our studies of deep and shallow rule learning, we define rule-based theories in a networked structure, which we describe in the following. We build upon the shallow two-level networks which we have previously used for experimenting with mini-batch rule learning
\citep{fb:DeCoDeML-20}, but generalize them from a shallow DNF-structure to deeper networks.

\subsection{Network Structure}
%\citet{fb:DeCoDeML-20} present a network whose output should be structured like a rule set in DNF. This network 
A conventional rule set consisting of multiple conjuctive rules that define a single target concept, corresponds to a logical expression in disjunctive normal form (DNF). The corresponding
% Such a 
network consists of three layers, the input layer, one hidden layer (=~\texttt{AND}~layer) and the output layer (=~\texttt{OR}~layer), as, e.g., illustrated in Figure~\ref{fig:parity-network}a. The input layer receives one-hot-encoded nominal attributes as binary features (=~literals), the hidden layer conjuncts these literals to rules and the output layer disjuncts the rules to a rule set. The network is designed for binary classification problems and produces a single prediction output that is \texttt{true} if and only if an input sample is covered by any of the rules in the rule set. 

For generalizing this structure to deeper networks, we need to define multiple layers.
Since the number of input features does not change and only a single output is predicted, the input layer and the output layer remain the same size. However, the number and the size of the hidden layers can be chosen arbitrarily. Note that we can still emulate a shallow DNF-structure by choosing a single hidden layer. In the more general case, the hidden layers are treated alternately as conjunctive and disjunctive layers. We focus on layer structures starting with a conjunctive hidden layer and ending with a disjunctive output layer, i.e. networks with an odd number of hidden layers. In this way, the output will be easier to compare with rule sets in DNF. Furthermore, the closer we are to the output layer, the more extensive are the rules and rule sets, and the smaller is the chance to form new meaningful combinations from them. As a consequence, the number of nodes per hidden layer should be lower the closer it is to the output layer. This makes the network shaped like a funnel.

\subsection{Network Weights and Initialization}
In the following, we assume the network to have $n+2$ layers, with each layer $i$ containing $s_i$ nodes. Layer $0$ corresponds to the input layer with $s_0=|\mathbf{x}|$ and layer $n+1$ to the output layer with $s_{n+1}=1$.
%of the hidden layers $i$, the network is built and the weights are initialized. 
Furthermore, a weight $w^{(i)}_{jk}$ is identified by the layer $i$ it belongs to, the node $j$ from which it receives the output, and the node $k$ in the successive layer $i+1$ to which it passes the activation. Thus, the weights of each layer can be represented by an $s_i \times s_{i-1}$-dimensional matrix $W^{(i)} = [ w^{(i)}_{jk} ] $.
In total, there are \(\sum_{i=0}^{n} s_i s_{i+1}\) Boolean weights which have to be learned, i.e., have to be set to \texttt{true} (resp. $1$) or \texttt{false} (resp.\ $0$). If weight $w^{(i)}_{jk}$ is set to \texttt{true}, this means that the output of node $j$ is used in the conjunction (if $i \bmod 2 = 0$) or disjunction (if $i \bmod 2 = 1$) that defines node $k$. If it is set to \texttt{false}, this output is ignored by node $k$.

In the beginning, these weights need to be initialized.
This initialization process is influenced by two hyperparameters: average rule length ($\bar{l}$) and initialization probability ($p$), where $\bar{l}$ only affects the weights in the first layer. Here we use the additional information which literals belong to the same attribute to avoid immediate contradictions within the first conjunction. Let $|\mathcal{A}|$ be the number of attributes, then each attribute is selected with the probability $\bar{l}/|\mathcal{A}|$ so that on average for $\bar{l}$ literals of different attributes the corresponding weight will be set to \texttt{true}. In the remaining layers, the weights are set to \texttt{true} with the probability $p$. Additionally, at least one outgoing weight from each node will be set to \texttt{true} to ensure connectivity. This implies that, regardless of the choice of $p$, all the weights in the last layer will always be initialized with \texttt{true} because there is only one output node. Note that, as a consequence, shallow DNF-structured networks will not be influenced by the choice of $p$, since they only consist of the first layer influenced by $\bar{l}$ and the last layer initialized with \texttt{true}.

\subsection{Prediction}

The prediction of the network can be efficiently computed using binary matrix multiplications ($\odot$).
%When the initialization is finished, the network can provide a prediction for samples at any time. 
In each layer $i$, the input features $A^{(i)}$ are multiplied with the corresponding weights $W^{(i)}$ and aggregated at the receiving node in layer $i+1$. If the aggregation is disjunctive, this directly corresponds to a binary matrix multiplication. According to De Morgan's law, 
%\begin{equation}
$
    a \land b = \neg ( \neg a \lor \neg b )
$
%\end{equation}
holds. This means that binary matrix multiplication can be used also in the conjunctive case, provided that the inputs and outputs are negated before and after the multiplication. Because of the alternating sequence of conjunctive and disjunctive layers, binary matrix multiplications and negations are also always alternated when passing data through the network, so that a binary matrix multiplication followed by a negation can be considered as a \texttt{NOR}-node. Thus, the activations $A^{(i+1)}$ can be computed from the activations in the previous layers as
\begin{equation}
    A^{(i+1)} \longleftarrow \tilde{A}^{(i)} \odot W^{(i)} 
\label{eq:activation}
\end{equation}
where $\tilde{X} = J - X$ denotes the element-wise negation of a matrix $X$ ($J$ denotes a matrix of all ones).
Hence, internally, we do not distinguish between conjunctive and disjunctive layers within the network, but have a uniform network structure consisting only of \texttt{NOR}-nodes. However, for the sake of the ease of interpretation, we chose to represent the networks as alternating \texttt{AND} and \texttt{OR} layers.

In the first layer, we have the choice whether to start with a disjunctive layer or a conjunctive one, which can be controlled by simply using the original input vector ($A^{(0)} = \mathbf{x}$) or its negation ($A^{(0)} = \tilde{\mathbf{x}}$) as the first layer.
Also, 
%The behavior of this network can only be customized by an additional negation before the \texttt{NOR}-node,
if the last layer is conjunctive, an additional negation must be performed at the end of the network so that the output has the same ''polarity'' as the target values. 
%, but this negation results automatically from the number of layers and the type of the first layer and is therefore not freely eligible.
%
In our experiments, we always start with a conjunctive and end with a disjunctive layer. In this way, the rule networks can be directly converted into conjunctive rule sets.

\begin{comment}
Equation~\ref{eq:activation} shows the computation of the activations $A^{(i)}$ using $I$ as the original input features in the first layer and $\sim$ to mark negations of activations. 

\begin{equation}
\begin{matrix}
  A^{(i)}=\begin{cases}
    \tilde{I},                         & \text{if $i=0$ and first layer conjunctive}.\\
    I,                                 & \text{if $i=0$ and first layer disjunctive}.\\
    \tilde{A}^{(i-1)} \odot W^{(i-1)}, & \text{otherwise}.
  \end{cases} \\
  A^{(n)}=\tilde{A}^{(n)} \hspace{9.5mm}\text{if last layer conjunctive}
\end{matrix}
\label{eq:activation}
\end{equation}
\end{comment}

\subsection{Training}

Following \citet{fb:DeCoDeML-20}, we implement a straight-forward mini-batch based greedy optimization scheme.
While the number, the arrangement and the aggregation types of the nodes remain unchanged, the training process will flip the weights of the network to optimize its outcome. %, as usual in neural networks.
Flipping a weight from $0$ to $1$ (or vice versa) can be understood to be a single addition (or removal) of a literal to the conjunction or disjunction encoded by the following node. A detailed pseudo-code of the  training process is shown in Algorithms~\ref{alg:fit}~and~\ref{alg:opt}. Given a deep rule network classifier, abbreviated as $drnc$, training samples $x$, their correct targets $y$ and an appropriate batch size for the training set, Algorithm~\ref{alg:fit} shows a na\text{\"i}ve greedy approach to fit the network to the training data. 

\begin{table}
\begin{minipage}{0.5\linewidth}
\begin{algorithm}[H]
\small
\SetKwInput{KwInput}{Input}
\SetKwInput{KwOutput}{Output}
\SetAlgoLined
\KwInput{drnc, $x$, $y$, batch\_size}
\KwOutput{drnc}
    drnc.initialize()\;
    n\_samples $\gets$ length($x$)\;
    n\_batches $\gets$ n\_samples / batch\_size\;
    best\_accuracy $\gets$ accuracy\_score($y$, drnc.predict($x$))\;
    best\_coefs $\gets$ drnc.coefs\;
    \For{$e\gets$0 \KwTo n\_epochs $- 1$}{
        $x$, $y\gets$ shuffle($x$, $y$)\;
        \For{$b\gets0$ \KwTo n\_batches $- 1$}{
            $x_{mb}$, $y_{mb}$ $\gets$ next\_batch(batch\_size)\;
            coefs $\gets$ drnc.optimize\_coefs($x_{mb}$, $y_{mb}$)\;
            accuracy $\gets$ accuracy\_score($y$, drnc.predict($x$))\;
            \If{accuracy $>$ best\_accuracy}{
                best\_accuracy $\gets$ accuracy\;
                best\_coefs $\gets$ drnc.coefs\;
            }
        }
    }
    drnc.coefs $\gets$ best\_coefs\;
    drnc.optimize\_coefs($x$, $y$)\;
    \Return{drnc}
    \caption{Deep Rule Network Training, \texttt{fit()}-method}
    \label{alg:fit}
\end{algorithm}
\end{minipage}
\begin{minipage}{0.49\linewidth}
\begin{algorithm}[H]
\small
\SetKwInput{KwInput}{Input}
\SetKwInput{KwOutput}{Output}
\SetAlgoLined
\KwInput{drnc, $x$, $y$, best\_accuracy}
\KwOutput{coefs}
    best\_$i\gets i$\;
    best\_$j\gets j$\;
    best\_$k\gets k$\;
    optimal $\gets$ $false$\;
    flip\_count $\gets 0$\;
    \While{$\neg$ optimal $\land$ flip\_count $<$ drnc.max\_flips}{
        \For{$i\gets$0 \KwTo n\_layers $- 1$}{
            \For{$j\gets0$ \KwTo n\_nodes[$i$] $- 1$}{
                \For{$k\gets$0 \KwTo n\_nodes[$i - 1$] $- 1$}{
                    flip(drnc.coefs[$i$][$j$][$k$])\;
                    accuracy $\gets$ accuracy\_score($y$, drnc.predict($x$))\;
                    \If{accuracy $>$ best\_accuracy}{
                        best\_accuracy $\gets$ accuracy\;
                        best\_coefs $\gets$ drnc.coefs\;
                        best\_$i\gets i$\;
                        best\_$j\gets j$\;
                        best\_$k\gets k$\;
                    }
                    flip(drnc.coefs[$i$][$j$][$k$])\;
                }
            }
        }
        \If{$\neg$ optimal}{
            flip(drnc.coefs[$i$][$j$][$k$])\;
            best\_accuracy $\gets$ accuracy\;
            best\_coefs $\gets$ drnc.coefs\;
            flip\_count $\gets$ flip\_count $+ 1$\;
        }
    }
    \Return{coefs}
    \caption{Deep Rule Network Training, \texttt{optimize\_coefs()}-method}
    \label{alg:opt}
\end{algorithm}
\end{minipage}
\end{table}

After the initialization, the base accuracy on the complete training set and the initial weights are stored as $\mathit{best\_accuracy}$ and $\mathit{best\_coefs}$ respectively. These values are updated every time when the predicted accuracy on the training set exceeds the previous maximum after processing a mini-batch of training examples. However, the predictive performance does not necessarily increase continuously, since the accuracy is optimized not on the whole training set, but on a mini-batch which is passed to Algorithm~\ref{alg:opt}. For all layers and nodes, the flips are executed and evaluated, whereas for flips in the first layer, weights of literals of the same attribute might be flipped as well, to respect the constraints of the one-hot-encoding of the input attributes. Finally, the flip with the biggest improvement of the accuracy on the current mini-batch is selected. These greedy adjustments are repeated until either no flip improves the accuracy on the mini-batch or a number of $\mathit{max\_flips}$ is reached, which ensures that the network does not overfit to the mini-batch data.

When all mini-batches are processed, the procedure is repeated until a fixed number of epochs is completed. Only the composition of the mini-batches is changed in each epoch by shuffling the training data before proceeding. After all epochs, the weights of the networks are reset to $\mathit{best\_coefs}$, the optimum found so far, and a final optimization on the complete training set is conducted to eliminate any overfitting on mini-batches. The returned network can then be used to predict outcomes of any further test instances.

\section{Experiments}
\label{sec:experiments}

In this section we present the results of differently structured rule networks on both artificial and real-world UCI datasets, with the goal of investigating the effect of differences in the depth of the networks. We first describe the artificial datasets (Section~\ref{sec:artificial-datasets}), then some preliminary experiments that helped us to focus on suitable network structures and hyperparameters (Section~\ref{sec:hyperparameter-grid-search}), and finally discuss the main results on the artificial and real datasets.

\subsection{Artificial Datasets}
\label{sec:artificial-datasets}

In Section~\ref{sec:motivation}, we presented the \texttt{XOR} or parity problem as a suitable dataset to compare the performance of shallow and deep rule networks. However, it has the disadvantage that the intermediate concepts do not have any information gain because the target class is determined only by the attributes which are missing in this concept. Thus, an artificial dataset suitable for our greedy optimization algorithm should not only include intermediate concepts which are meaningful but also a strictly monotonically decreasing entropy between these concepts, so that they can be learned in a stepwise fashion in successive layers. One way to generate artificial datasets that satisfy these requirements is to take the output of a randomly generated deep rule network. Subsequently, this training information can be used to see whether the function encoded in the original network can be recovered. Note that such a recovery is also possible for networks with different layer structures. In particular, each of the logical functions encoded in such a deep network can, of course, also be encoded as a DNF expression, so that shallow networks are not in an a priori disadvantage (provided that their hidden layer is large enough, which we ensure in preliminary experiments reported in Section~\ref{sec:hyperparameter-grid-search}).

We use a dataset of ten Boolean inputs named $a$ to $j$ and generate all possible $2^{10}$ combinations as training or test samples. These samples are extended by the ten negations $\neg a$ to $\neg j$ via one-hot-encoding and finally passed to a funnel-shaped deep rule network with $n=5$ and $\mathbf{s}=[32, 16, 8, 4, 2]$. The weights of the network are set by randomly initializing the network and then training it on two randomly selected examples, one assigned to the positive and one to the negative class, to ensure both a positive and negative output is possible. 
%The weights of the network are set by randomly initializing the network and then greedily optimizing it so two randomly selected examples, one positive and one negative, are correctly predicted by the network.
%n the first sample to the negative class, the second one to the positive class and fit the network on those two samples. 
If the resulting ratio of positively predicted samples is still less then $20\%$ or more than $80\%$, the network is reinitialized with a new random seed to avoid extremely imbalanced datasets.

\begin{figure}[t]
\begin{minipage}[b]{0.38\textwidth}
	\centering
    \includegraphics[height=7.5cm]{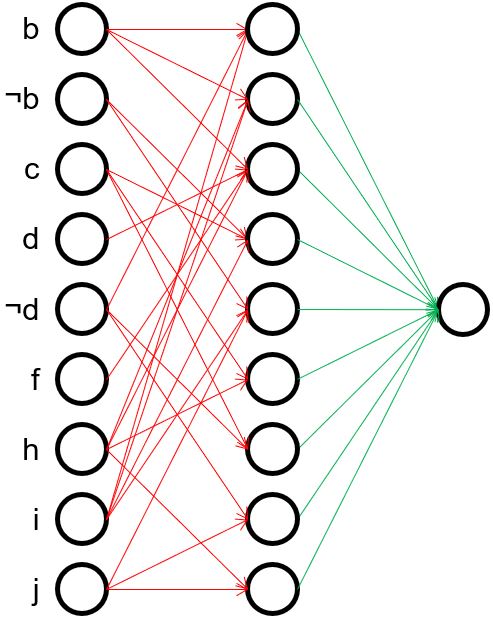}
    \medskip
    a) shallow example
\end{minipage}%
\begin{minipage}[b]{0.61\textwidth}
	\centering
    \includegraphics[height=7.5cm]{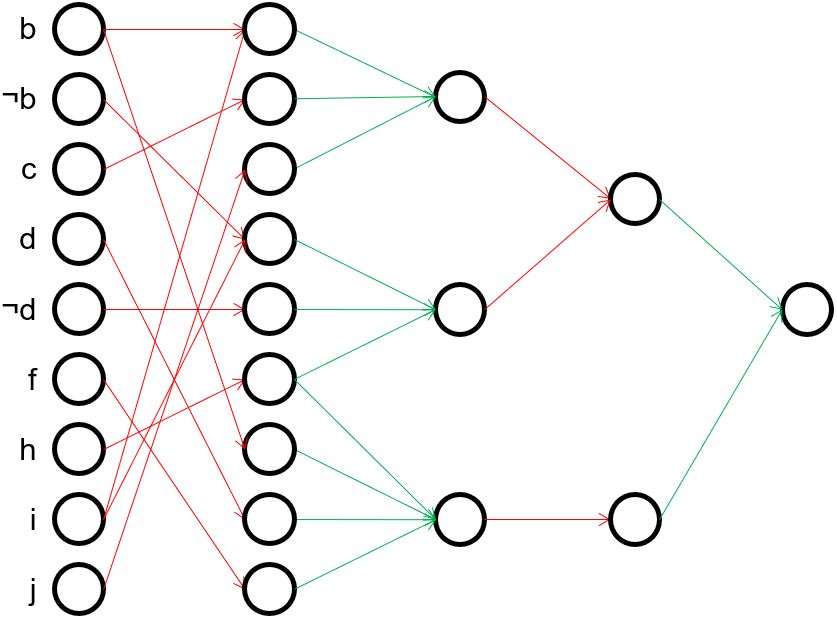}
    \medskip
    b) deep example
\end{minipage}
\caption{Example networks of Equations~\ref{eq:shallow}~and~\ref{eq:deep}. For a better overview, the input nodes not influencing the final concept are removed. Red connections are logical ANDs, green edges correspond to logical ORs.}
\label{fig:deep_shallow_example}
\end{figure}

An example concept for a generated dataset is shown in Figure~\ref{fig:deep_shallow_example}. Thinking of a rule network, circles represent nodes and are connected by an arrow if and only if the corresponding weight is \texttt{true}. Note that many nodes and weights are irrelevant, and are not shown, e.g., neither $a$ nor $\neg a$ have an influence on the generated output.
The flat representation shown in Figure~\ref{fig:deep_shallow_example}a corresponds to the following DNF expression\footnote{We used the python library \textsl{Sympy} (\url{https://www.sympy.org/en/index.html}) to compute the corresponding DNF representation.}:
\begin{equation}
(b \land \neg d \land i) \lor (b \land h \land i) \lor (b \land d \land f \land h) \lor (\neg b \land c \land i) \lor (\neg b \land i \land j) \lor (c \land h) \lor (c \land \neg d) \lor (\neg d \land j) \lor (h \land j)
\label{eq:shallow}
\end{equation}
It can be clearly seen that
the rules in this simplified formula share some common features, indicating intermediate concepts in subsequent layers which are combined in the end. A more compact representation of the same concept using a hierarchical structure is:
\begin{equation}
\biggl(\Bigl((b \land i) \lor c \lor j\Bigr) \land \Bigl((\neg b \land i) \lor \neg d \lor h\Bigr)\biggr) \lor \biggl(b \land d \land f \land h\biggr)
\label{eq:deep}
\end{equation}
The second representation only needs $11$ aggregations ($6$ \texttt{AND}, $5$ \texttt{OR}) in comparison to $23$ aggregations ($15$ \texttt{AND}, $8$ \texttt{OR}) in the first one. This is also reflected in the number of weights set to \texttt{true}, i.e. the number of arrows in Figure~\ref{fig:deep_shallow_example} ($33$ in \ref{fig:deep_shallow_example}a vs. $26$ in \ref{fig:deep_shallow_example}b). In contrast, when training the deep rule network, we must learn at least $180 + 27 + 6 + 2 = 215$ binary weights correctly, while for the shallow one already $180 + 9 = 189$ would be sufficient. Note that we included here the eleven nodes in the input layer which are absent in the formulas, but whose weights nevertheless have to be learned by both networks. Looking at the figures, it is also noticeable that even though both networks have sparse weight matrices, the ones of the hierarchical network are even more sparse which makes it almost shaped like a tree. In the following experiments, we evaluate which of these two representations is easier to learn approximately.

\subsection{Hyperparameter Tuning}
\label{sec:hyperparameter-grid-search}

Before the main experiments, we conducted a few preliminary experiments on three of the artificial datasets to set suitable default values for the hyperparameters of the networks. 
One of these hyperparameters also known from neural networks is the number of epochs ($\mathit{n\_epochs}$). By definition (Algorithm~\ref{alg:fit}), the accuracy monotonically increases with a higher number of epochs while at the same time the training time rises as well. After five epochs, the performance no longer rises remarkably, so this value seems as a good trade-off between performance and training time.

The second hyperparameter $\mathit{batch\_size}$ affects these two measures as well. After tests using the number of instances as $\mathit{batch\_size}$ and others skipping the final optimization in Algorithm~\ref{alg:fit} on the full batch, we notice that using a combination of mini-batches and full batches performs better than either of the two individual batch variants. For the artificial datasets, a $\textit{batch\_size}$ of 50 is suitable. Finally, the limitation of iterations per mini-batch by $\textit{max\_flips}$ also influences both the accuracy and training time. However, in case of noise-free artificial data, we can leave $\mathit{max\_flips}$ unbounded to achieve the optimal performance.

The experiments also showed no clear advantages or disadvantages between a conjunctive or disjunctive first layer, so in the following experiments we focus on networks starting with a conjunctive layer which offer the biggest similarity and best comparability to models learned from classic rule learners. Furthermore, we dispense with a separate optimization of the last layer like in \cite{fb:DeCoDeML-20}, as this did not result in any improvement in performance.

For the remaining hyperparameters, we tried to find appropriate settings by performing a grid search on 20 artificial datasets. The hyperparameters to be optimized are the average rule length $\bar{l}$, the initialization probability $p$, the number $n$ and the sizes $s_i$ of hidden layers.
The other hyperparameters are set to the default values stated above, except that only a single epoch is used in order to speed up the grid search. This will have a negative effect on the performance in general, but should not significantly change the  ranking of the different networks. 

\begingroup
\setlength{\tabcolsep}{6pt} % Default value: 6pt
\renewcommand{\arraystretch}{1.2} % Default value: 1
\begin{table}[tbph]
\caption{Hyperparameters for deep and shallow networks}
\centering
\begin{tabular}{|l||l|l|}
\hline
          & deep                                      & shallow                      \\
\hline\hline
          & 5: [72, 36, 12, 6, 2], [32, 16, 8, 4, 2], &                              \\
$n, s_i$  & 4: [36, 12, 6, 2], [16, 8, 4, 2],         & 1: 10, 20, 50, 100, 200, 500 \\
          & 3: [12, 6, 2], [8, 4, 2]                  &                              \\
\hline
$\bar{l}$ & 1, 2, 3                                   & 1, 2, 3, 4, 5, 6, 7          \\
\hline
$p$       & 0.025, 0.075, 0.125                       & -                            \\
\hline
\end{tabular}
\label{tab:grid_search}
\end{table}
\endgroup

%The following values are tested in the grid search 
Table~\ref{tab:grid_search} shows the hyperparameters that we compared in a grid search.
For the deep networks, we set $n$ to values from $3$ to $5$. On the one hand, this guarantees that they contain at least two conjunctive and two disjunctive layers to map a wide variety of hierarchical concepts effectively. On the other hand, it still permits that the values of $s_i$ can be set to values bigger than $10$ while maintaining a reasonable training time with the na\text{\"i}ve greedy algorithm. We create two different networks for each of the three values of $n$ and set the values $s_i$ so that $s_i \geq 2s_{i+1}$, resulting in a smaller network and a bigger one containing $1.5$ to $3.5$ times as many weights that can be adapted. For shallow networks, $n$ is by definition set to $1$. To ensure that these networks have approximately the same expressive power as the corresponding deep networks, we set $s_1$ so that the total number of weights in both network types is roughly the same. Additionally, we try a very high number of $s_1=500$ rules to estimate if a very big single layer can improve the accuracy remarkably. For the average initialized rule length $\bar{l}$ we will use the integer values from $1$ to $3$ for deep networks and from $1$ to $7$ for shallow ones. We assume that in shallow networks a higher value of $\bar{l}$ is required, while in deep networks the intermediate concepts can be combined in successive layers. The numerical deficit of deep network test cases caused by $\bar{l}$ is compensated by the additional hyperparameter $p$, where we use three values between $2.5\%$ and $12.5\%$. Therefore, in total, the accuracy of the deep network will be mapped to three dimensions $n|s_i$, $\bar{l}$ and $p$ and the accuracy of the shallow network to only two dimensions $s_1$ and $\bar{l}$. 

\begin{figure}[!htbp]
\centering
\begin{tabular}{cccc}
\includegraphics[height=5.7cm]{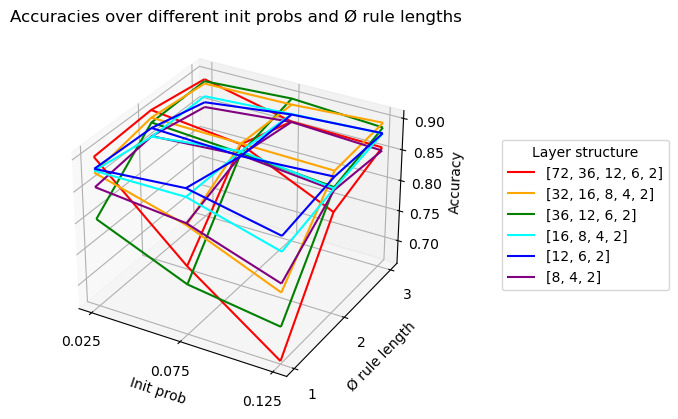} \\
(a)  \\[6pt]
\end{tabular}
\begin{tabular}{cccc}
\includegraphics[height=5.7cm]{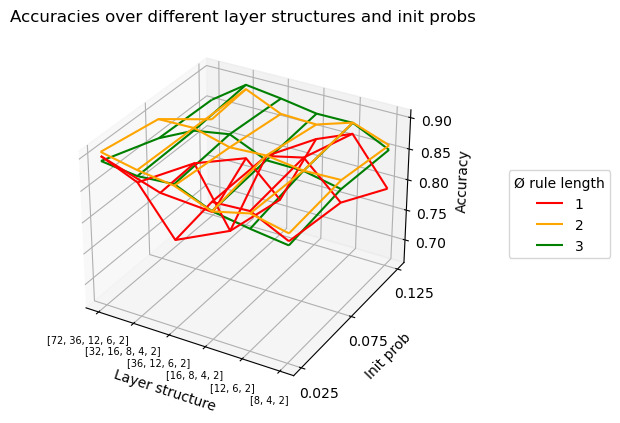} &
\includegraphics[height=5.7cm]{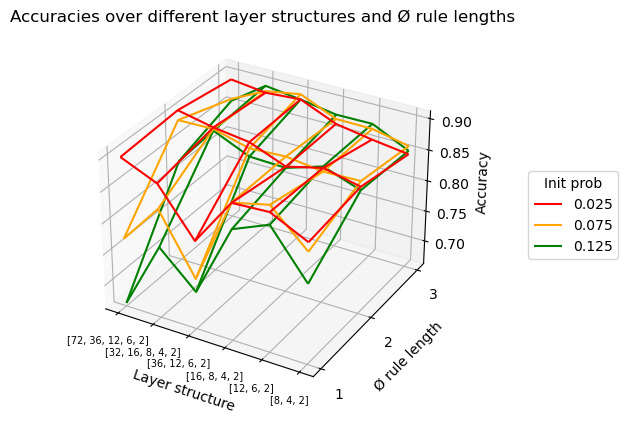} \\
(b) & (c)  \\[6pt]
\end{tabular}
\begin{tabular}{cccc}
\includegraphics[height=5.7cm]{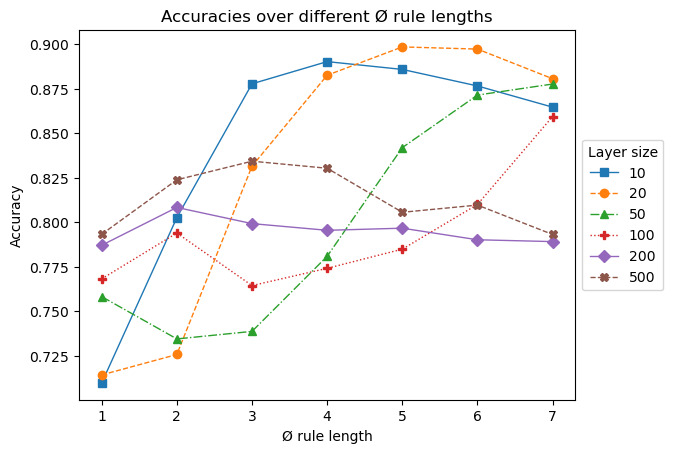} &
\includegraphics[height=5.7cm]{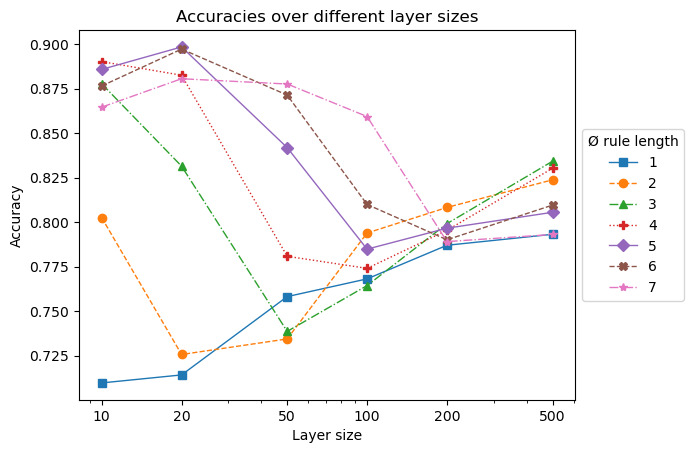} \\
(d) & (e)  \\[6pt]
\end{tabular}
\caption{Grid search on the hyperparameters average rule length $\bar{l}$, initialization probability $p$, number $n$ and sizes $s_i$ of hidden layers. The first three plots show accuracies for deep rule networks with different combinations of \textbf{(a)} $p$ and $\bar{l}$, \textbf{(b)} $n|s_i$ and $\bar{l}$ and \textbf{(c)} $n|s_i$ and $p$. The last two show accuracies for shallow rule networks with different \textbf{(d)} $\bar{l}$ and \textbf{(e)} $s_1$.}
\label{fig:grid_search}
\end{figure}

The results of the grid search are shown in Figure~\ref{fig:grid_search}.
The optimal hyperparameter setting for deep networks with an accuracy of $0.9073$ is $\mathbf{s}=[72, 36, 12, 6, 2]$, $\bar{l}=2$ and $p=0.025$. However, we notice in Figure~\ref{fig:grid_search}a that the red curve of the largest structure $\mathbf{s}=[72, 36, 12, 6, 2]$ not only contains the maximum, but also the minimum accuracy with $\bar{l}=1$ and $p=0.125$. While in general the combination of a lower $\bar{l}$ and a higher $p$ decreases the accuracy, this effect seems to be stronger the bigger the network structure is. Despite the higher sensitivity, the larger layer structures provide better maximum accuracies than the smaller ones, as can be seen in the upper left corner of the graph. When comparing the graphs for different values of $\bar{l}$ in Figure~\ref{fig:grid_search}b, it is noticeable that, with only few exceptions, the red curve for $\bar{l}=1$ lies below the other two curves. The same can be observed in Figure~\ref{fig:grid_search}c, here for the green curve for $p=0.125$. Combinations of other values of $\bar{l}$ and $p$ provide good accuracies regardless of the layer structure.

For shallow networks, we can see a high correlation between $s_1$ and $\bar{l}$ in Figure~\ref{fig:grid_search}d. The graphs (except of the red one) resemble a downward-opening parabola, so that the accuracy becomes lower and lower the greater the deviation from this optimal value. Thereby applies that the smaller the layer size, the larger is the optimal value of $\bar{l}$, e.g. $5$ for $20$ and $2$ for $200$. Finally, Figure~\ref{fig:grid_search}e shows that, contrary to the the deep networks, the sensitivity to $\bar{l}$ decreases with the size of the shallow network. The optimal accuracy of $0.8984$ is reached when the shallow network hyperparameters are set to $s_1=20$ and $\bar{l}=5$.

Based on the above results, we will only use three network versions for
the main experiments reported in the following sections. As a candidate for shallow networks, we take the best combination of $s_1=20$ and $\bar{l}=5$. For the deep networks, however, we will choose the second best network $\mathbf{s}=[32, 16, 8, 4, 2]$ combined with $\bar{l}=2$ and an averaged $p=0.05$, since it is almost ten times faster than the best deep network while still reaching an accuracy over $0.895$. The third network is chosen as an intermediate stage between the first two: $\mathbf{s}=[32, 8, 2]$ combined with $\bar{l}=3$ and $p=0.05$. While still being a deep network, the learned rules can be passed to the output layer a little faster.

\subsection{Results on Artificial Datasets}

In these experiments, we use a combination of $15$ artificial datasets with seeds we already used in the hyperparameter grid search and $5$ artificial datasets with new seeds to detect potential overfitting on the first datasets. 
Note that even if the network structure $\mathbf{s}=[32, 16, 8, 4, 2]$ of the generating network would allow to create more complex formulas, we have seen in Figure~\ref{fig:deep_shallow_example} that its actual generated target concept can often be reproduced by a smaller network. In particular, we ensured for all of the generated datasets that the DNF concept does not contain more than $20$ rules, so that it can be theoretically also be learned by the tested shallow network with $s_1=20$ (and therefore also for the two deep networks, since their first layer is already bigger).

All datasets are tested with the networks presented at the end of subsection~\ref{sec:hyperparameter-grid-search}, using five epochs, a batch size of $50$ and an unlimited number of flips per batch. 
In the following figures and tables, we will refer to these networks based on their number of layers, i.e. \drnc{5} for $\mathbf{s}=[32, 16, 8, 4, 2]$, \drnc{3} for $\mathbf{s}=[32, 8, 2]$ and \rnc\ for $s_1=20$.
For computational reasons, all of the reported results were estimated with a simple 1$\times$2-fold cross validation. While this may not yield the most reliable estimate on each individual dataset, we nevertheless get a coherent picture over all 20 datasets, as we will see in the following. 

\begin{figure}[b]
\centering
\includegraphics[height=7.8cm]{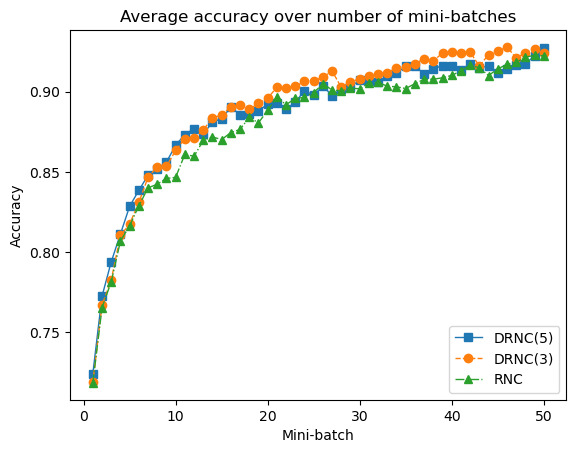}
\caption{Average accuracy of rule network with 1/3/5 layers on training dataset.}
\label{fig:avg_artificial}
\end{figure}

Figure~\ref{fig:avg_artificial} shows the accuracies on the training set averaged on all $20$ datasets. On the x-axis, the number of processed mini-batches is shown, whereby after every ten mini-batches a new epoch starts. The base accuracy before processing the first mini-batch and after the full batch optimization are omitted. We can see that the deep networks not only deliver higher accuracies but they also converge slightly faster than the shallow one. The orange curve of \drnc{3} runs a little higher than the blue one of \drnc{5}, whereas the green curve of \rnc\ has some distance to them, especially during the first two epochs.

Table~\ref{tab:artificial} shows the accuracies of the three networks. 
For each dataset, the best accuracy of the three network classifiers is highlighted in bold. We can see a clear advantage for the two deep networks both when considering the average accuracy and the amount of highest accuracies. 
The results clearly show that the best performing deep networks outperform the best performing shallow network in all but 4 of the 20 generated datasets. Both the average rank and the average accuracy of the deep networks is considerably better than the corresponding values for \rnc. This also holds for pairwise comparisons of the columns (\drnc{5} vs.\ \rnc{} 15:5, \drnc{3} vs.\ \rnc\  15:5).
%
%For the average rank, we only take the three network approaches into account. 
The Friedman-test for the ranks yields a significance of more than $95\%$. A subsequent Nemenyi-Test delivers a critical distance of 0.741 ($95\%$) or 0.649 ($90\%$), which shows that \drnc{3} and \rnc{} are significantly different on a level of more than $95\%$ and \drnc{5} and \rnc\ on a level of more than $90\%$. We thus find it safe to conclude that deep networks outperform shallow networks on these datasets.

\begingroup
\setlength{\tabcolsep}{6pt} % Default value: 6pt
\renewcommand{\arraystretch}{1.2} % Default value: 1
\begin{table}[H]
\caption{Accuracies on artificial datasets. Rule network with 1/3/5 layers vs \ripper vs \CART.}
\label{tab:artificial}
\sisetup{detect-weight,mode=text}
% use \B for bold numbers
\renewrobustcmd{\bfseries}{\fontseries{b}\selectfont}
\renewrobustcmd{\boldmath}{}
\newrobustcmd{\B}{\bfseries}
\medskip
\centering
\begin{tabular}{|lS|SSS|SS|}
\hline
seed          & {\%(+)}        & \drnc{5}      & \drnc{3}     & \rnc           & \ripper & \CART    \\
\hline\hline
5             & 0.4453         & 0.958           & \B 0.9863       & 0.9531          & 0.9805   & 0.9844    \\
16            & 0.7959         & 0.9639          & \B 0.9707       & 0.9629          & 0.9766   & 0.9551    \\
19            & 0.6562         & \B 1            & 0.9902          & 0.9746          & 1        & 1         \\
24            & 0.584          & 0.9053          & 0.9043          & \B 0.916        & 0.9463   & 0.9404    \\
36            & 0.6943         & 0.8828          & \B 0.9209       & 0.9043          & 0.8867   & 0.9111    \\
44            & 0.7939         & \B 0.9629       & 0.9551          & 0.9326          & 0.9482   & 0.9697    \\
53            & 0.6055         & \B 0.9805       & \B 0.9805       & 0.9775          & 0.9746   & 0.9824    \\
57            & 0.7705         & \B 0.9824       & 0.9736          & 0.9639          & 0.9951   & 0.9902    \\
60            & 0.7715         & 0.9443          & \B 0.9453       & 0.9209          & 0.958    & 0.9883    \\
65            & 0.5312         & \B 0.9854       & 0.9688          & 0.9414          & 0.9961   & 0.9922    \\
68            & 0.5654         & 0.9248          & 0.9443          & \B 0.9619       & 0.9688   & 0.9355    \\
69            & 0.6924         & 0.9551          & \B 0.9658       & 0.9199          & 0.9795   & 0.9717    \\
70            & 0.6338         & 0.9014          & 0.9062          & \B 0.9229       & 0.9111   & 0.8984    \\
81            & 0.5684         & 0.9004          & \B 0.9131       & 0.8857          & 0.9248   & 0.9756    \\
82            & 0.7188         & 0.9941          & \B 0.998        & 0.9717          & 1        & 1         \\
85            & 0.5312         & \B 1            & 0.998           & 0.9736          & 1        & 1         \\
89            & 0.6084         & 0.8926          & 0.9434          & \B 0.9629       & 0.9502   & 0.9395    \\
107           & 0.6172         & \B 0.8965       & 0.873           & 0.8643          & 0.9043   & 0.9277    \\
112           & 0.7549         & \B 0.9346       & 0.9248          & 0.9189          & 0.9082   & 0.9561    \\
118           & 0.5957         & \B 0.9688       & 0.9414          & 0.9434          & 0.9736   & 0.9688    \\
\hline
\multicolumn{2}{|l|}{Ø Accuracy} & 0.9467          & 0.9502          & 0.9386          & 0.9591   & 0.9644    \\
\multicolumn{2}{|l|}{Ø Rank}     & 1.775           & 1.725           & 2.5             &          &           \\
\hline
\end{tabular}
\end{table}
\endgroup

In the two right-most columns of 
Table~\ref{tab:artificial} we also show a comparison to the state-of-the-art rule learner \ripper\ \citep{Ripper} and the decision tree learner \CART\ \citep{CART} in Python implementations using default parameters.\footnote{We used the implementations available from \url{https://pypi.org/project/wittgenstein/} and \url{https://scikit-learn.org/stable/modules/generated/sklearn.tree.DecisionTreeClassifier.html}.}
We see that all network approaches are outperformed by the \ripper\ and \CART\ classifiers with default setting. The difference between \ripper\ and \drnc{3} is approximately the same as the difference between \drnc{3} and \rnc. However, considering that we only use a na\"ive greedy algorithm, it could not be expected (and was also not our objective) to be able to beat state-of-the-art rule learner. These results also confirm that shallow rule learners (of which both \ripper\ and \CART\ are representatives) had no disadvantage by the way we generated the datasets.

\subsection{Results on UCI Datasets}
For an estimation how the rule networks perform on real-world datasets, we select the six classification datasets \textsl{car-evaluation}, \textsl{connect-4}, \textsl{kr-vs-kp}, \textsl{mushroom}, \textsl{tic-tac-toe} and \textsl{vote} from the UCI Repository \citep{Dua:2019}. They differ in the number of attributes and instances, but have in common that they consist only of nominal attributes. However, the datasets \textsl{car-evaluation} and \textsl{connect-4} are actually multi-class classification problems and are therefore converted into the binary classification problem whether a sample belongs to the most frequent class or not. Of all six binary classification problems, the networks to be tested treat again the more common class as the positive class and the less common as the negative class. As with the artificial datasets, we additionally compare the performance of the networks to \ripper\ and \CART, and again all accuracies are obtained via 1$\times$2-fold cross validation. 
% \emph{Note: For both deep networks in the first run one dataset led to a performance on the majority baseline level. We therefore use a reinitialization of the network with a different seed if the training accuracy does not exceed the ratio of positive samples. In order not to create an unfair advantage for the two deep networks thereby, we also allow the shallow network to use two different seeds and take the best accuracy achieved.}
In case a random initialization did not yield any result (i.e., the resulting network classified all examples into a single class), we re-initialized with a different seed (this happened once for both deep network versions).
The results are shown in Table~\ref{tab:real_world}.

\begingroup
\setlength{\tabcolsep}{6pt} % Default value: 6pt
\renewcommand{\arraystretch}{1.2} % Default value: 1
\begin{table}[H]
\caption{Accuracies on real-world datasets. Rule network with 1/3/5 layers vs \texttt{RIPPER} vs \texttt{CART}.}
\label{tab:real_world}
\sisetup{detect-weight,mode=text}
% use \B for bold numbers
\renewrobustcmd{\bfseries}{\fontseries{b}\selectfont}
\renewrobustcmd{\boldmath}{}
\newrobustcmd{\B}{\bfseries}
\medskip
\centering
\begin{tabular}{|lS|SSS|SS|}
\hline
dataset        & {\%(+)}   & \drnc{5}  & \drnc{3}  & \rnc      & \ripper & \CART \\
\hline \hline
car-evaluation & 0.7002    & 0.8999     & \B 0.9022  & 0.8565     & 0.9838   & 0.9821 \\
connect-4      & 0.6565    & \B 0.7728  & 0.7712     & 0.7597     & 0.7328   & 0.8195 \\
kr-vs-kp       & 0.5222    & 0.9671     & 0.9643     & \B 0.9725  & 0.9812   & 0.989  \\
mushroom       & 0.784     & \B 1       & 0.978      & 0.993      & 0.9992   & 1      \\
tic-tac-toe    & 0.6534    & 0.8956     & 0.9196     & \B 0.9541  & 1        & 0.9217 \\
vote           & 0.6138    & \B 0.9655  & 0.9288     & 0.9264     & 0.8804   & 0.9287 \\
\hline
\multicolumn{2}{|l|}{Ø Rank} & 1.667      & 2.167      & 2.167      &        &          \\
\hline
\end{tabular}
\end{table}
\endgroup

We can again observe that the deep 5-layer network \drnc{5} outperforms the shallow network \rnc{}.  Of all rule networks, it provides the highest accuracy on the \textsl{connect-4}, \textsl{mushroom} and \textsl{vote} datasets, whereas \texttt{DRNC(3)} performs best on \textsl{car-evaluation} and \texttt{RNC} on \textsl{kr-vs-kp} and \textsl{tic-tac-toe}. 
The latter two datasets are also interesting:  \textsl{tic-tac-toe} clearly does not require a deep structure, because for solving it, the learner essentially needs to enumerate all three-in-a-row positions on a $3 \times 3$ board. 
This is similar to \textsl{connect-4}, where four-in-a-row positions have to be recognized. However, in the former case,
%is particularly interesting. Even though both games are about getting a certain number of tiles connected in a row, column or diagonal, they also have two crucial differences. Firstly, in \textsl{connect-4} four stones have to be combined instead of just three, and second, the playing field there is also larger than in \textsl{tic-tac-toe}. This leads to the fact that in the latter 
there is only one matching tile for an intermediate concept consisting of two tiles, while in \textsl{connect-4} there are several, which can potentially be exploited by a deeper network. In the 
\textsl{kr-vs-kp} dataset, deep structures are also not helpful because it consists of carefully engineered features for the KRKP chess endgame, which were designed in an iterative process so that the game can be learned with a decision tree learner \citep{lig*Shapiro87}. It would be an ambitious goal of deep rule learning methods to be able to learn such a dataset from, e.g., only the positions of the chess pieces. This is clearly beyond the state-of-the-art of current rule learning algorithms.

The comparison to \ripper\ and \CART\ is again clearly in favor of these state-of-the-art algorithms. Interestingly, \ripper\ performs rather badly on the \textsl{connect-4} and \textsl{vote} datasets, which we speculate is due to some bug in the Python implementation that we used. Conversely, the rule networks perform extremely bad on the \textsl{car-evaluation} dataset.

%Between the rule networks we can not identify a big difference in the average rank anymore, even if \texttt{DRNC(5)} seems to have the best performance.

\section{Conclusion}
\label{sec:conclusion}

We proposed a technique how deep and shallow rule systems can be learned by using a network approach with a greedy optimization algorithm. For both deep and shallow rule networks, we find good hyperparameter settings that allow the networks to reach reasonable accuracies on both artificial and real-world datasets, even though the approach is still outperformed by state-of-the-art learning algorithms such as \ripper\ and \CART. Our experiments on both artificial and real-world benchmark data indicate that deep rule networks outperform shallow networks. The deep networks obtain not only a higher accuracy, but also need less mini-batch iterations to achieve it. Moreover, the experiments in the hyperparameter grid search indicate that the deep networks are generally more robust to the choice of the hyperparameters than shallow networks. On the other hand, we also had some cases on real-world data sets where deep networks failed because a poor initialization resulted in indiscriminate predictions. 

\section{Future Work}
\label{sec:future-work}

In this work, it was not our goal to reach a state-of-the-art predictive performance, but instead we wanted to evaluate a very simple greedy optimization algorithm on both shallow and deep networks, in order to get an indication on the potential of deep rule networks. Nevertheless, several avenues for improving our networks have surfaced, which we intend to explore in the near future.

One of the main drawbacks of the presented deep rule networks is the extremely high runtime due to the primitive flipping algorithm. A single flip needs a recalculation of all activations in the network, even if only a few them will be affected by this flip whereby the matrix multiplication could be minimized considerably. Conversely, this knowledge can be used to find a small subset of flips that affects a certain activation. On the other hand, the majority of possible flips does not have any effect on this activation or the accuracy at all. This effect will typically remain unchanged after a few more flips are done. Therefore an exhaustive search of all flips is only needed in the first iteration, while afterwards just a subset of possible flips should be considered which can be built either in a deterministic or probabilistic way.

Due to this lack of backpropagation, the flips are evaluated by their influence on the prediction when executed. However, when looking at a false positive, we can only correct this error by making the overall hypothesis of the network more specific. In order to achieve this, only flips from \texttt{false} to \texttt{true} in conjunctive layers or flips from \texttt{true} to \texttt{false} in disjunctive layers have to be taken into account, and vice versa, to achieve a generalization of the hypothesis. This way all flips are split into two groups "generalization-flips" and "specification-flips" of which only one group has to be considered at the same time. This improvement as well as the above mentioned selection of a subset of flips might also allow us to perform two or more flips at the same time so that a better result than with the greedy approach can be achieved.

An even more promising approach starts one step earlier in the initialization phase of the network. Instead of specifying the structure of the network and finding optimal initialization parameters $\bar{l}$ and $p$ for it, a small part of the data could be used to create a rough draft version of the network. The Quine-McCluskey algorithm \citep{Quine/McCluskey} or \ripper are suitable methods to generate shallow networks, whereas the \texttt{ESPRESSO}-algorithm \citep{DBLP:books/sp/BraytonHMS84} would generate deep networks. Decision trees can also be used to generate deep networks since the contained rules already share some conditions and, moreover, similar subtrees can be merged. In the case of decision trees, the network can also be expanded easily to multi-class classification.

All these approaches share some significant advantages over the network approach we developed so far. First of all, the decision which class value will be treated as positive or negative does not have to be made manually any longer. Second, they automatically deliver a suitable initialization of the network, which otherwise would have to be improved by similar approaches like used in neural networks \citep[e.g.,][]{DBLP:conf/ccwc/RamosNY17} to achieve a robust performance. Third, the general structure of the network is not limited to a fixed size and depth where each node is strictly assigned to a specific layer. Instead of generating nodes that become useless after a few flips have been processed and that should be removed, we can thereby start with a small structure which can be adapted purposefully by copying and mutating good nodes and pruning bad ones. However, it remains unclear whether these changes still lead to improvements in performance or if the network in the given structure is already optimal.

\section*{Acknowledgments}
%This is a short text to acknowledge the contributions of specific colleagues, institutions, or agencies that aided the efforts of the authors.
We are grateful to Eneldo Loza Menc\'{i}a, Michael Rapp and Eyke H\"ullermeier for inspiring discussions, fruitful pointers to related work, and for suggesting the evaluation with artificial datasets.

%\section*{Data Availability Statement}
%The datasets [GENERATED/ANALYZED] for this study can be found in the [NAME OF REPOSITORY] [LINK].
% Please see the availability of data guidelines for more information, at https://www.frontiersin.org/about/author-guidelines#AvailabilityofData

\bibliographystyle{unsrtnat}
\bibliography{bib/bibliography,bib/rules,bib/interpretability,bib/ilp,bib/ml,bib/nn,bib/ensembles,bib/jf,bib/multilabel}

\begin{thebibliography}{56}
\providecommand{\natexlab}[1]{#1}
\providecommand{\url}[1]{\texttt{#1}}
\expandafter\ifx\csname urlstyle\endcsname\relax
  \providecommand{\doi}[1]{doi: #1}\else
  \providecommand{\doi}{doi: \begingroup \urlstyle{rm}\Url}\fi

\bibitem[Michalski(1969)]{AQ}
Ryszard~S. Michalski.
\newblock On the quasi-minimal solution of the covering problem.
\newblock In \emph{Proceedings of the 5th International Symposium on
  Information Processing (FCIP-69)}, volume A3 (Switching Circuits), pages
  125--128, Bled, Yugoslavia, 1969.

\bibitem[F{\"{u}}rnkranz et~al.(2012)F{\"{u}}rnkranz, Gamberger, and Lavra{\v
  c}]{jf:Book-Nada}
Johannes F{\"{u}}rnkranz, Dragan Gamberger, and Nada Lavra{\v c}.
\newblock \emph{Foundations of Rule Learning}.
\newblock Springer-Verlag, 2012.
\newblock ISBN 978-3-540-75196-0.

\bibitem[Wang et~al.(2017)Wang, Rudin, Doshi{-}Velez, Liu, Klampfl, and
  MacNeille]{RuleSets-Bayesian}
Tong Wang, Cynthia Rudin, Finale Doshi{-}Velez, Yimin Liu, Erica Klampfl, and
  Perry MacNeille.
\newblock A {B}ayesian framework for learning rule sets for interpretable
  classification.
\newblock \emph{Journal of Machine Learning Research}, 18:\penalty0
  70:1--70:37, 2017.
\newblock URL \url{http://jmlr.org/papers/v18/16-003.html}.

\bibitem[Lakkaraju et~al.(2016)Lakkaraju, Bach, and
  Leskovec]{InterpretableDecisionSets}
Himabindu Lakkaraju, Stephen~H. Bach, and Jure Leskovec.
\newblock Interpretable decision sets: {A} joint framework for description and
  prediction.
\newblock In Balaji Krishnapuram, Mohak Shah, Alexander~J. Smola, Charu~C.
  Aggarwal, Dou Shen, and Rajeev Rastogi, editors, \emph{Proceedings of the
  22nd {ACM} {SIGKDD} International Conference on Knowledge Discovery and Data
  Mining (KDD-16)}, pages 1675--1684, San Francisco, CA, 2016. {ACM}.
\newblock \doi{10.1145/2939672.2939874}.
\newblock URL \url{http://doi.acm.org/10.1145/2939672.2939874}.

\bibitem[Cohen(1995)]{Ripper}
William~W. Cohen.
\newblock Fast effective rule induction.
\newblock In A.~Prieditis and S.~Russell, editors, \emph{Proceedings of the
  12th International Conference on Machine Learning (ML-95)}, pages 115--123,
  Lake Tahoe, CA, 1995. Morgan Kaufmann.

\bibitem[Hornik(1991)]{NN-UniversalApproximation}
Kurt Hornik.
\newblock Approximation capabilities of multilayer feedforward networks.
\newblock \emph{Neural Networks}, 4\penalty0 (2):\penalty0 251 -- 257, 1991.
\newblock ISSN 0893-6080.
\newblock \doi{https://doi.org/10.1016/0893-6080(91)90009-T}.
\newblock URL
  \url{http://www.sciencedirect.com/science/article/pii/089360809190009T}.

\bibitem[Mhaskar et~al.(2017)Mhaskar, Liao, and Poggio]{NN-WhyDeep}
Hrushikesh Mhaskar, Qianli Liao, and Tomaso~A. Poggio.
\newblock When and why are deep networks better than shallow ones?
\newblock In Satinder~P. Singh and Shaul Markovitch, editors, \emph{Proceedings
  of the 31st {AAAI} Conference on Artificial Intelligence}, pages 2343--2349,
  San Francisco, California, {USA}, 2017. {AAAI} Press.
\newblock URL \url{http://aaai.org/ocs/index.php/AAAI/AAAI17/paper/view/14849}.

\bibitem[Friedman and Popescu(2008)]{RuleFit}
Jerome~H. Friedman and Bogdan~E. Popescu.
\newblock Predictive learning via rule ensembles.
\newblock \emph{The Annals of Applied Statistics}, 2\penalty0 (3):\penalty0
  916--–954, 2008.
\newblock \doi{10.1214/07-AOAS148}.

\bibitem[Breiman(2001)]{RandomForests}
Leo Breiman.
\newblock Random forests.
\newblock \emph{Machine Learning}, 45\penalty0 (1):\penalty0 5--32, 2001.

\bibitem[Dembczy\'nski et~al.(2010)Dembczy\'nski, Kot{\l}owski, and
  S{\l}owiński]{ENDER}
Krzysztof Dembczy\'nski, Wojciech Kot{\l}owski, and Roman S{\l}owiński.
\newblock {ENDER:} a statistical framework for boosting decision rules.
\newblock \emph{Data Mining and Knowledge Discovery}, 21\penalty0 (1):\penalty0
  52--90, 2010.
\newblock \doi{10.1007/s10618-010-0177-7}.
\newblock URL \url{https://doi.org/10.1007/s10618-010-0177-7}.

\bibitem[Rapp et~al.(2020)Rapp, Loza~Menc{\'\i}a, F{\"u}rnkranz, Nguyen, and
  H{\"u}llermeier]{mr:ECML-PKDD-20}
Michael Rapp, Eneldo Loza~Menc{\'\i}a, Johannes F{\"u}rnkranz, Vu-Linh Nguyen,
  and Eyke H{\"u}llermeier.
\newblock Learning gradient boosted multi-label classification rules.
\newblock In Frank Hutter, Kristian Kersting, Jefrey Lijffijt, and Isabel
  Valera, editors, \emph{Proceedings of the European Conference on Machine
  Learning and Knowledge Discovery in Databases (ECML/PKDD), Part III}, volume
  12459 of \emph{Lecture Notes in Computer Science}, pages 124--140.
  Springer-Verlag, 2020.
\newblock URL \url{http://arxiv.org/abs/2006.13346}.

\bibitem[Angelino et~al.(2017)Angelino, Larus{-}Stone, Alabi, Seltzer, and
  Rudin]{RuleLists-Optimal}
Elaine Angelino, Nicholas Larus{-}Stone, Daniel Alabi, Margo~I. Seltzer, and
  Cynthia Rudin.
\newblock Learning certifiably optimal rule lists for categorical data.
\newblock \emph{Journal of Machine Learning Research}, 18:\penalty0
  234:1--234:78, 2017.
\newblock URL \url{http://jmlr.org/papers/v18/17-716.html}.

\bibitem[F{\"u}rnkranz et~al.(2020)F{\"u}rnkranz, H{\"u}llermeier,
  Loza~Menc{\'i}a, and Rapp]{jf:DeCoDeML-20-Challenge}
Johannes F{\"u}rnkranz, Eyke H{\"u}llermeier, Eneldo Loza~Menc{\'i}a, and
  Michael Rapp.
\newblock Learning structured declarative rule sets -- a challenge for deep
  discrete learning.
\newblock In Kristian Kersting, Stefan Kramer, and Zahra Ahmadi, editors,
  \emph{Proceedings of the 2nd Workshop on Deep Continuous-Discrete Machine
  Learning (DeCoDeML)}, 2020.

\bibitem[F{\"{u}}rnkranz(2005)]{jf:Dagstuhl-04}
Johannes F{\"{u}}rnkranz.
\newblock From local to global patterns: Evaluation issues in rule learning
  algorithms.
\newblock In K.~Morik, Jean-Fran{\c c}ois Boulicaut, and Arno Siebes, editors,
  \emph{Local Pattern Detection}, pages 20--38. Springer-Verlag, 2005.

\bibitem[Matheus(1989)]{CI-Framework}
Christopher~J. Matheus.
\newblock A constructive induction framework.
\newblock In \emph{Proceedings of the 6th International Workshop on Machine
  Learning}, pages 474--475, 1989.

\bibitem[Stahl(1996)]{ILP-PI}
Irene Stahl.
\newblock Predicate invention in {Inductive Logic Programming}.
\newblock In L.~De~Raedt, editor, \emph{Advances in Inductive Logic
  Programming}, volume~32 of \emph{Frontiers in Artificial Intelligence and
  Applications}, pages 34--47. IOS Press, 1996.

\bibitem[Wnek and Michalski(1994)]{AQ17-HCI}
Janusz Wnek and Ryszard~S. Michalski.
\newblock Hypothesis-driven constructive induction in {AQ17-HCI}: A method and
  experiments.
\newblock \emph{Machine Learning}, 14\penalty0 (2):\penalty0 139--168, 1994.
\newblock Special Issue on Evaluating and Changing Representation.

\bibitem[Pfahringer(1994)]{CiPF}
Bernhard Pfahringer.
\newblock Controlling constructive induction in {CiPF}: an {MDL} approach.
\newblock In Pavel~B. Brazdil, editor, \emph{Proceedings of the 7th European
  Conference on Machine Learning (ECML-94)}, Lecture Notes in Artificial
  Intelligence, pages 242--256, Catania, Sicily, 1994. Springer-Verlag.

\bibitem[Muggleton(1987)]{Duce}
Stephen~H. Muggleton.
\newblock Structuring knowledge by asking questions.
\newblock In Ivan Bratko and N.~Lavra\v{c}, editors, \emph{Progress in Machine
  Learning}, pages 218--229. Sigma Press, Wilmslow, England, 1987.

\bibitem[Muggleton and Buntine(1988)]{CIGOL}
Stephen~H. Muggleton and Wray~L. Buntine.
\newblock Machine invention of first-order predicates by inverting resolution.
\newblock In \emph{Proceedings of the 5th International Conference on Machine
  Learning (ML-88)}, pages 339--352, 1988.

\bibitem[Kijsirikul et~al.(1992)Kijsirikul, Numao, and Shimura]{CHAMP}
Boonserm Kijsirikul, Masayuki Numao, and Masamichi Shimura.
\newblock Discrimination-based constructive induction of logic programs.
\newblock In \emph{Proceedings of the 10th National Conference on Artificial
  Intelligence (AAAI-92)}, pages 44--49, 1992.

\bibitem[Kok and Domingos(2007)]{PI-Statistical}
Stanley Kok and Pedro~M. Domingos.
\newblock Statistical predicate invention.
\newblock In Zoubin Ghahramani, editor, \emph{Proceedings of the 24th
  International Conference on Machine Learning {(ICML}-07)}, volume 227 of
  \emph{{ACM} International Conference Proceeding Series}, pages 433--440,
  Corvallis, Oregon, USA, 2007. {ACM}.

\bibitem[Morik et~al.(1993)Morik, Wrobel, Kietz, and Emde]{Mobal}
Katharina Morik, Stefan Wrobel, J{\"o}rg-Uwe Kietz, and Werner Emde.
\newblock \emph{Knowledge Acquisition and Machine Learning -- Theory, Methods,
  and Applications}.
\newblock Academic Press, London, 1993.

\bibitem[Sommer(1996)]{Sommer-Diss}
Edgar Sommer.
\newblock \emph{Theory Restructuring -- A Perspective on Design and Maintenance
  of Knowlege Based Systems}, volume 171 of \emph{{DISKI}}.
\newblock Infix, 1996.

\bibitem[Muggleton et~al.(2015)Muggleton, Lin, and
  Tamaddoni{-}Nezhad]{ILP-PI-MetaInterpretative}
Stephen~H. Muggleton, Dianhuan Lin, and Alireza Tamaddoni{-}Nezhad.
\newblock Meta-interpretive learning of higher-order dyadic datalog:
  {P}redicate invention revisited.
\newblock \emph{Machine Learning}, 100\penalty0 (1):\penalty0 49--73, 2015.
\newblock \doi{10.1007/s10994-014-5471-y}.
\newblock URL \url{https://doi.org/10.1007/s10994-014-5471-y}.

\bibitem[Kramer(2020)]{HigherLevelRepresentations}
Stefan Kramer.
\newblock A brief history of learning symbolic higher-level representations
  from data (and a curious look forward).
\newblock In \emph{Proceedings of the 29th International Joint Conference on
  Artificial Intelligence (IJCAI), Survey Track}, pages 4868--4876, 2020.

\bibitem[Malerba et~al.(1997)Malerba, Semeraro, and
  Esposito]{LabelDepRuleLearning}
D.~Malerba, G.~Semeraro, and F.~Esposito.
\newblock A multistrategy approach to learning multiple dependent concepts.
\newblock In G.~Nakhaeizadeh and C.~C. Taylor, editors, \emph{Machine Learning
  and Statistics: The Interface}, chapter~4, pages 87--106. Wiley, London,
  England, 1997.

\bibitem[De~Raedt et~al.(1993)De~Raedt, Lavra\v{c}, and D\v{z}eroski]{MPL}
Luc De~Raedt, Nada Lavra\v{c}, and Sa\v{s}o D\v{z}eroski.
\newblock Multiple predicate learning.
\newblock In R.~Bajcsy, editor, \emph{Proceedings of the 13th International
  Joint Conference on Artificial Intelligence (IJCAI-93)}, pages 1037--1043,
  Chamb\'{e}ry, France, 1993. Morgan Kaufmann.

\bibitem[Tsoumakas and Katakis(2007)]{Multilabel-Overview}
Grigorios Tsoumakas and Ioannis Katakis.
\newblock Multi-label classification: An overview.
\newblock \emph{International Journal of Data Warehousing and Mining},
  3\penalty0 (3):\penalty0 1--17, 2007.

\bibitem[Tsoumakas et~al.(2010)Tsoumakas, Katakis, and
  Vlahavas]{tsoumakas10MLoverview}
Grigorios Tsoumakas, Ioannis Katakis, and Ioannis~P. Vlahavas.
\newblock Mining multi-label data.
\newblock In Oded Maimon and Lior Rokach, editors, \emph{Data Mining and
  Knowledge Discovery Handbook}, pages 667--685. Springer, 2nd edition, 2010.
\newblock ISBN 978-0-387-09823-4.
\newblock \doi{10.1007/978-0-387-09823-4_34}.
\newblock URL \url{http://lpis.csd.auth.gr/publications/tsoumakas09-dmkdh.pdf}.

\bibitem[Zhang and Zhou(2014)]{zhang2014review}
Min{-}Ling Zhang and Zhi{-}Hua Zhou.
\newblock A review on multi-label learning algorithms.
\newblock \emph{{IEEE} Transactions on Knowledge and Data Engineering},
  26\penalty0 (8):\penalty0 1819--1837, 2014.

\bibitem[Waegeman et~al.(2019)Waegeman, Dembczy{\'{n}}ski, and
  H{\"u}llermeier]{MultiTargetPrediction}
Willem Waegeman, Krzysztof Dembczy{\'{n}}ski, and Eyke H{\"u}llermeier.
\newblock Multi-target prediction: a unifying view on problems and methods.
\newblock \emph{Data Mining and Knowledge Discovery}, 33\penalty0 (2):\penalty0
  293--324, 2019.

\bibitem[Dembczy\'nski et~al.(2012)Dembczy\'nski, Waegeman, Cheng, and
  H{\"u}llermeier]{MultiLabel-Dependence}
Krzysztof Dembczy\'nski, Willem Waegeman, Weiwei Cheng, and Eyke
  H{\"u}llermeier.
\newblock On label dependence and loss minimization in multi-label
  classification.
\newblock \emph{Machine Learning}, 88\penalty0 (1-2):\penalty0 5--45, 2012.

\bibitem[Read et~al.(2011)Read, Pfahringer, Holmes, and
  Frank]{ClassifierChains}
Jesse Read, Bernhard Pfahringer, Geoff Holmes, and Eibe Frank.
\newblock Classifier chains for multi-label classification.
\newblock \emph{Machine Learning}, 85\penalty0 (3):\penalty0 333--359, 2011.

\bibitem[Read et~al.(2021)Read, Pfahringer, Holmes, and
  Frank]{ClassifierChains-Perspective}
Jesse Read, Bernhard Pfahringer, Geoff Holmes, and Eibe Frank.
\newblock Classifier chains: {A} review and perspectives.
\newblock \emph{Journal of Artificial Intelligence Research}, 70:\penalty0
  683--718, 2021.
\newblock \doi{10.1613/jair.1.12376}.
\newblock URL \url{https://doi.org/10.1613/jair.1.12376}.

\bibitem[Burkhardt and Kramer(2015)]{Burkhardt2015}
Sophie Burkhardt and Stefan Kramer.
\newblock On the spectrum between binary relevance and classifier chains in m
  ulti-label classification.
\newblock In Roger~L. Wainwright, Juan~Manuel Corchado, Alessio Bechini, and
  Jiman Hong, editors, \emph{Proceedings of the 30th Annual {ACM} Symposium on
  Applied Computing (SAC)}, pages 885--892, Salamanca, Spain, 2015. {ACM}.

\bibitem[Read and Hollm{\'{e}}n(2014)]{Read-Hollmen-IDA-14}
Jesse Read and Jaakko Hollm{\'{e}}n.
\newblock A deep interpretation of classifier chains.
\newblock In Hendrik Blockeel, Matthijs van Leeuwen, and Veronica Vinciotti,
  editors, \emph{Advances in Intelligent Data Analysis 13 (IDA)}, volume 8819
  of \emph{Lecture Notes in Computer Science}, pages 251--262, Leuven, Belgium,
  2014. Springer.
\newblock \doi{10.1007/978-3-319-12571-8\_22}.

\bibitem[Read and Hollm{\'{e}}n(2015)]{Read-Hollmen-arxiv-15}
Jesse Read and Jaakko Hollm{\'{e}}n.
\newblock Multi-label classification using labels as hidden nodes.
\newblock \emph{CoRR}, abs/1503.09022, 2015.
\newblock URL \url{http://arxiv.org/abs/1503.09022}.

\bibitem[Nam et~al.(2016)Nam, Loza~Menc{\'{\i}}a, and
  F{\"{u}}rnkranz]{jn:AAAI-16}
Jinseok Nam, Eneldo Loza~Menc{\'{\i}}a, and Johannes F{\"{u}}rnkranz.
\newblock All-in text: Learning document, label, and word representations
  jointly.
\newblock In Dale Schuurmans and Michael~P. Wellman, editors, \emph{Proceedings
  of the 30th AAAI Conference on Artificial Intelligence}, pages 1948--1954.
  AAAI Press, 2016.

\bibitem[H{\"{u}}llermeier et~al.(2020)H{\"{u}}llermeier, F{\"{u}}rnkranz,
  Menc{\'{\i}}a, Nguyen, and Rapp]{eh:RuleML-20}
Eyke H{\"{u}}llermeier, Johannes F{\"{u}}rnkranz, Eneldo~Loza Menc{\'{\i}}a,
  Vu{-}Linh Nguyen, and Michael Rapp.
\newblock Rule-based multi-label classification: Challenges and opportunities.
\newblock In V{\'{\i}}ctor Guti{\'{e}}rrez{-}Basulto, Tom{\'{a}}s Kliegr, Ahmet
  Soylu, Martin Giese, and Dumitru Roman, editors, \emph{Proceedings of the 4th
  International Joint Conference on Rules and Reasoning (RuleML+RR)}, volume
  12173 of \emph{Lecture Notes in Computer Science}, pages 3--19, Oslo, Norway,
  2020. Springer.
\newblock \doi{10.1007/978-3-030-57977-7\_1}.
\newblock URL \url{https://doi.org/10.1007/978-3-030-57977-7\_1}.

\bibitem[Poon and Domingos(2011)]{SPNs}
Hoifung Poon and Pedro~M. Domingos.
\newblock Sum-product networks: {A} new deep architecture.
\newblock In F{\'{a}}bio~Gagliardi Cozman and Avi Pfeffer, editors,
  \emph{Proceedings of the 27th Conference on Uncertainty in Artificial
  Intelligence (UAI)}, pages 337--346, Barcelona, Spain, 2011. {AUAI} Press.
\newblock URL
  \url{https://dslpitt.org/uai/displayArticleDetails.jsp?mmnu=1\&smnu=2\&article\_id=2194\&proceeding\_id=27}.

\bibitem[Delalleau and Bengio(2011)]{SPN-DeepVsShallow}
Olivier Delalleau and Yoshua Bengio.
\newblock Shallow vs. deep sum-product networks.
\newblock In John Shawe{-}Taylor, Richard~S. Zemel, Peter~L. Bartlett, Fernando
  C.~N. Pereira, and Kilian~Q. Weinberger, editors, \emph{Advances in Neural
  Information Processing Systems 24 (NIPS)}, pages 666--674, Granada, Spain,
  2011.
\newblock URL
  \url{https://proceedings.neurips.cc/paper/2011/hash/8e6b42f1644ecb1327dc03ab345e618b-Abstract.html}.

\bibitem[Cohen et~al.(2020)Cohen, Yang, and Mazaitis]{TensorLog}
William~W. Cohen, Fan Yang, and Kathryn Mazaitis.
\newblock {TensorLog}: {A} probabilistic database implemented using
  deep-learning infrastructure.
\newblock \emph{Journal of Artificial Intelligence Research}, 67:\penalty0
  285--325, 2020.
\newblock \doi{10.1613/jair.1.11944}.
\newblock URL \url{https://doi.org/10.1613/jair.1.11944}.

\bibitem[Evans and Grefenstette(2018)]{ILP-DeepNoisy}
Richard Evans and Edward Grefenstette.
\newblock Learning explanatory rules from noisy data.
\newblock \emph{Journal of Artificial Intelligence Research}, 61:\penalty0
  1--64, 2018.
\newblock \doi{10.1613/jair.5714}.
\newblock URL \url{https://doi.org/10.1613/jair.5714}.

\bibitem[Senge and H{\"{u}}llermeier(2011)]{FuzzyPatternTrees}
Robin Senge and Eyke H{\"{u}}llermeier.
\newblock Top-down induction of fuzzy pattern trees.
\newblock \emph{{IEEE} Transactions on Fuzzy Systems}, 19\penalty0
  (2):\penalty0 241--252, 2011.
\newblock \doi{10.1109/TFUZZ.2010.2093532}.
\newblock URL \url{https://doi.org/10.1109/TFUZZ.2010.2093532}.

\bibitem[Courbariaux et~al.(2015)Courbariaux, Bengio, and David]{BinaryConnect}
Matthieu Courbariaux, Yoshua Bengio, and Jean{-}Pierre David.
\newblock Binaryconnect: Training deep neural networks with binary weights
  during propagations.
\newblock In Corinna Cortes, Neil~D. Lawrence, Daniel~D. Lee, Masashi Sugiyama,
  and Roman Garnett, editors, \emph{Advances in Neural Information Processing
  Systems 28 (NIPS)}, pages 3123--3131, Montreal, Quebec, Canada, 2015.
\newblock URL
  \url{https://proceedings.neurips.cc/paper/2015/hash/3e15cc11f979ed25912dff5b0669f2cd-Abstract.html}.

\bibitem[Qin et~al.(2020)Qin, Gong, Liu, Bai, Song, and Sebe]{BinaryNN-Survey}
Haotong Qin, Ruihao Gong, Xianglong Liu, Xiao Bai, Jingkuan Song, and Nicu
  Sebe.
\newblock Binary neural networks: A survey.
\newblock \emph{Pattern Recognitition}, \penalty0 (105), 2020.
\newblock URL \url{https://arxiv.org/abs/2004.03333}.

\bibitem[Li et~al.(2016)Li, Zhang, and Liu]{TernaryNetworks}
Fengfu Li, Bo~Zhang, and Bin Liu.
\newblock Ternary weight networks.
\newblock \emph{arxiv}, abs/1605.04711, 2016.
\newblock URL \url{http://arxiv.org/abs/1605.04711}.

\bibitem[Zhu et~al.(2017)Zhu, Han, Mao, and
  Dally]{TernaryNetworks-Quantization}
Chenzhuo Zhu, Song Han, Huizi Mao, and William~J. Dally.
\newblock Trained ternary quantization.
\newblock In \emph{Proceedings of the 5th International Conference on Learning
  Representations (ICLR)}, Toulon, France, 2017. OpenReview.net.
\newblock URL \url{https://openreview.net/forum?id=S1\_pAu9xl}.

\bibitem[Beck and F{\"u}rnkranz(2020)]{fb:DeCoDeML-20}
Florian Beck and Johannes F{\"u}rnkranz.
\newblock An investigation into mini-batch rule learning.
\newblock In Kristian Kersting, Stefan Kramer, and Zahra Ahmadi, editors,
  \emph{Proceedings of the 2nd Workshop on Deep Continuous-Discrete Machine
  Learning (DeCoDeML)}, 2020.

\bibitem[Breiman et~al.(1984)Breiman, Friedman, Olshen, and Stone]{CART}
Leo Breiman, Jerome~H. Friedman, R.~Olshen, and C.~Stone.
\newblock \emph{Classification and Regression Trees}.
\newblock Wadsworth \& Brooks, Pacific Grove, CA, 1984.

\bibitem[Dua and Graff(2017)]{Dua:2019}
Dheeru Dua and Casey Graff.
\newblock {UCI} machine learning repository, 2017.
\newblock URL \url{http://archive.ics.uci.edu/ml}.

\bibitem[Shapiro(1987)]{lig*Shapiro87}
Alen~D. Shapiro.
\newblock \emph{Structured Induction in Expert Systems}.
\newblock Turing Institute Press. Addison-Wesley, 1987.

\bibitem[{McCluskey}(1956)]{Quine/McCluskey}
E.~J. {McCluskey}.
\newblock Minimization of {B}oolean functions.
\newblock \emph{The Bell System Technical Journal}, 35\penalty0 (6):\penalty0
  1417--1444, 1956.
\newblock \doi{10.1002/j.1538-7305.1956.tb03835.x}.

\bibitem[Brayton et~al.(1984)Brayton, Hachtel, McMullen, and
  Sangiovanni{-}Vincentelli]{DBLP:books/sp/BraytonHMS84}
Robert~K. Brayton, Gary~D. Hachtel, Curtis~T. McMullen, and Alberto~L.
  Sangiovanni{-}Vincentelli.
\newblock \emph{Logic Minimization Algorithms for {VLSI} Synthesis}, volume~2
  of \emph{The Kluwer International Series in Engineering and Computer
  Science}.
\newblock Springer, 1984.
\newblock ISBN 978-1-4612-9784-0.
\newblock \doi{10.1007/978-1-4613-2821-6}.
\newblock URL \url{https://doi.org/10.1007/978-1-4613-2821-6}.

\bibitem[Ramos et~al.(2017)Ramos, Nakakuni, and
  Yfantis]{DBLP:conf/ccwc/RamosNY17}
Ernesto~Zamora Ramos, Masanori Nakakuni, and Evangelos Yfantis.
\newblock Quantitative measures to evaluate neural network weight
  initialization strategies.
\newblock In \emph{{IEEE} 7th Annual Computing and Communication Workshop and
  Conference, {CCWC} 2017, Las Vegas, NV, USA, January 9-11, 2017}, pages 1--7.
  {IEEE}, 2017.
\newblock \doi{10.1109/CCWC.2017.7868389}.
\newblock URL \url{https://doi.org/10.1109/CCWC.2017.7868389}.

\end{thebibliography}

\end{document}